\documentclass{article}
\usepackage{ijcai23}

\usepackage{times}
\usepackage{soul}
\usepackage{url}
\usepackage[hidelinks]{hyperref}
\usepackage[utf8]{inputenc}
\usepackage[small]{caption}
\usepackage{graphicx}
\usepackage{amsmath}
\usepackage{amsthm}
\usepackage{booktabs}
\usepackage[switch]{lineno}

\usepackage{natbib} 
\usepackage{subfigure}
\usepackage{amsfonts}
\usepackage{gensymb}
\usepackage{multirow}
\usepackage{amssymb}
\usepackage{pifont}
\newcommand{\cmark}{\ding{51}}%
\newcommand{\xmark}{\ding{55}}%
\usepackage{booktabs, caption, tikz}
\usepackage{color, colortbl,stackengine,float,makecell}
\definecolor{Gray}{gray}{0.9}
\usepackage[ruled, vlined, linesnumbered]{algorithm2e}

\usepackage{stfloats, enumerate, enumitem}
\usepackage{lipsum}

\title{eTag: Class-Incremental Learning with Embedding Distillation and Task-Oriented Generation}

\author{
    Author Name
    \affiliations
    Affiliation
    \emails
    email@example.com
}

\author{
Libo Huang$^1$
\and
Yan Zeng$^2$\and
Chuanguang Yang$^{1}$\and
Zhulin An$^1$\and
Boyu Diao$^1$\And
Yongjun Xu$^1$
\affiliations
$^1$Institute of Computing Technology, Chinese Academy of Sciences\\
$^2$Tsinghua University\\
\emails
\{www.huanglibo, yanazeng013\}@gmail.com,
\{yangchuanguang, anzhulin, diaoboyu2012, xyj\}@ict.ac.cn
}

\begin{document}

\maketitle

\begin{abstract}
    Class-Incremental Learning (CIL) aims to solve the neural networks' catastrophic forgetting problem, which refers to the fact that once the network updates on a new task, its performance on previously-learned tasks drops dramatically. Most successful CIL methods incrementally train a feature extractor with the aid of stored exemplars, or estimate the feature distribution with the stored prototypes. However, the stored exemplars would violate the data privacy concerns, while the stored prototypes might not reasonably be consistent with a proper feature distribution, hindering the exploration of real-world CIL applications. In this paper, we propose a method of \textit{e}mbedding distillation and \textit{Ta}sk-oriented \textit{g}eneration (\textit{eTag}) for CIL, which requires neither the exemplar nor the prototype. Instead, eTag achieves a data-free manner to train the neural networks incrementally. To prevent the feature extractor from forgetting, eTag distills the embeddings of the network's intermediate blocks. Additionally, eTag enables a generative network to produce suitable features, fitting the needs of the top incremental classifier. Experimental results confirmed that our proposed eTag considerably outperforms the state-of-the-art methods on CIFAR-100 and ImageNet-sub\footnote{Our code will be available upon acceptance, thanks for your attention.}.
\end{abstract}

\section{Introduction}

Dynamic scenarios require the deployed model could manage the sequential arriving tasks~\citep{parisi2019continual}. On these tasks, models often incur the \textit{catastrophic forgetting} problem, which implies that the performance on the previously-learned tasks dramatically degenerates when the model learns a new one~\citep{mccloskey1989catastrophic}.

\begin{figure*}[!t]
  \centering
  \subfigure[DGR]{\includegraphics[height=3cm]{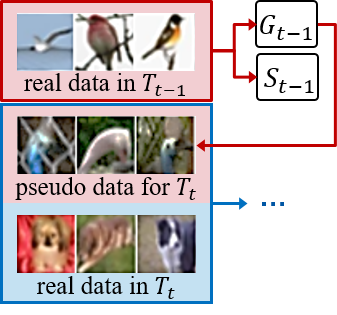}\label{fig:1a}}
  \hspace*{\fill}
  \tikz{\draw[densely dotted](0, 3) -- (0, 0);}
  \hspace*{\fill}
  \subfigure[Tag]{\includegraphics[height=3cm]{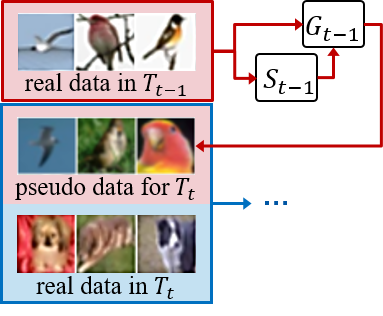}\label{fig:1b}}    
  \hspace*{\fill}
  \tikz{\draw[densely dotted](0, 3) -- (0,0);}
  \hspace*{\fill}
  \subfigure[eTag]{\includegraphics[height=3cm]{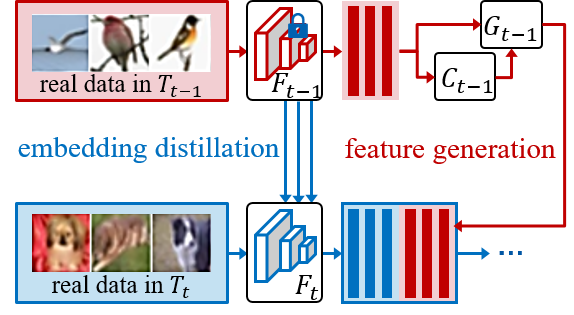}\label{fig:1c}}    
  \hspace*{\fill}
  \tikz{\draw[solid](0, 3) -- (0,0);}
  \hspace*{\fill}
  \subfigure[Preliminary results]{\includegraphics[height=3cm]{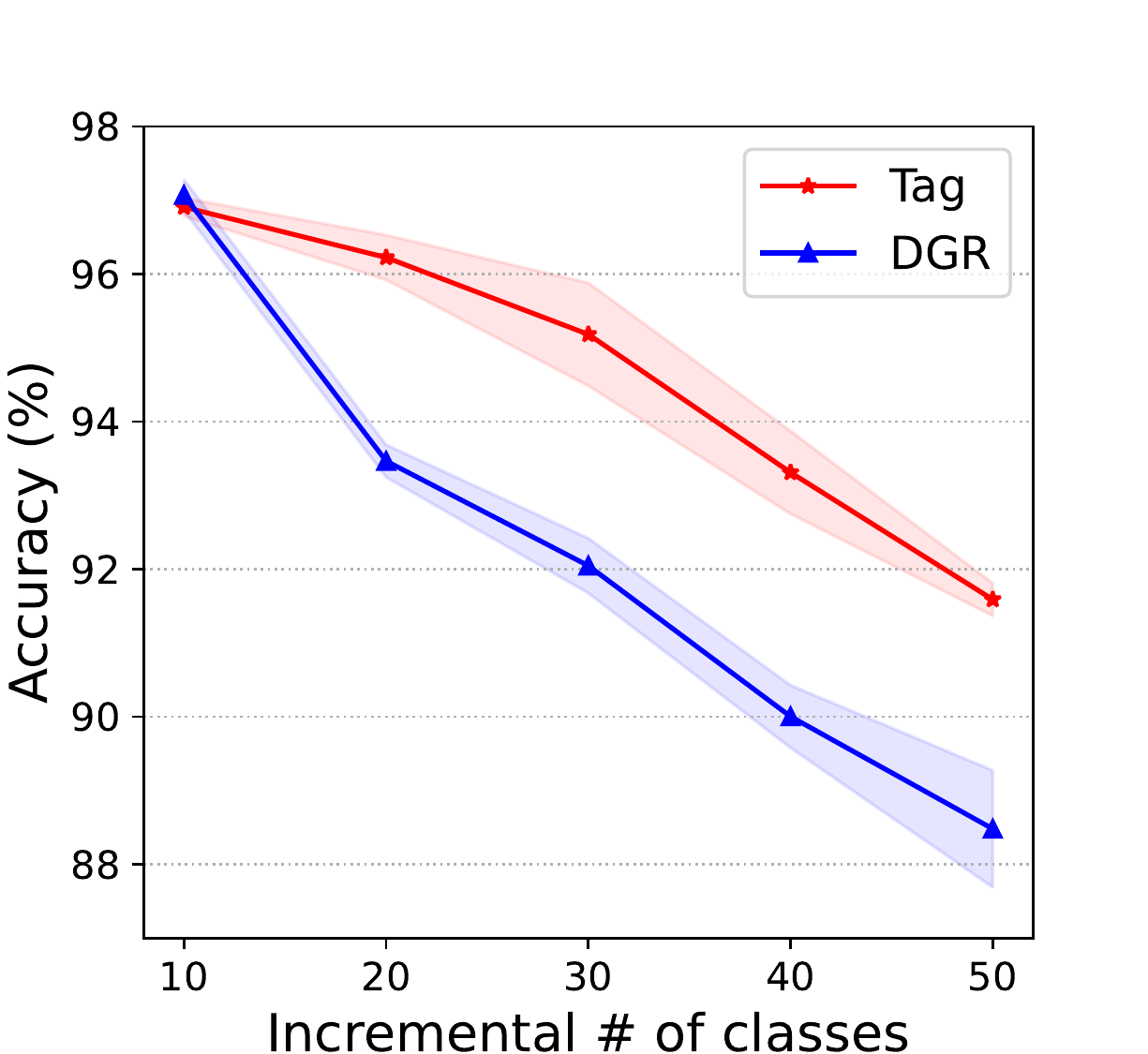}\label{fig:1d}}
  \caption{Comparison of DGR and our proposed Tag as well as the eTag. (a) DGR trains the generator $G_{t-1}$ on the task $T_{t-1}$ to generate the pseudo-samples when learning on task $T_t$. (b) Tag trains $G_{t-1}$ on $T_{t-1}$ with the guidance from trained and frozen solver $S_{t-1}$. (c) eTag split $S$ to feature extractor $F$ and the classifier $C$, which incrementally trains $F$ with embedding distillation and trains $C$ with Tag. (d) Preliminary results of generative sample replay methods, DGR and Tag, on five permutation MNIST CIL tasks.}
  \label{fig:1}
\end{figure*}

One of the most general solutions to the forgetting problem is Class-Incremental Learning (CIL)~\citep{aljundi2017expert,masana2020class}. The oracle of CIL assumes that the data of previously-learned tasks are always available, and thus combines such data with the current-task one to jointly train the CIL model (i.e., joint training), whereas this oracle violates the CIL setting where previous data are not allowed~\citep{chen2018lifelong,caruana1997multitask}. To achieve comparable performance, early CIL methods intuitively attempted to rebuild the training or inference environment of joint training. For instance\footnote{Methods of rebuilding inferences environment of joint training are as follows. Parameter isolation retains either a well-trained network~\citep{rusu2016progressive} or a set of parameters~\citep{li2017learning} for each learned task. The regularization-based method prevents the crucial parameters of learned tasks from adapting to the current one~\citep{kirkpatrick2017overcoming}.}, deep generative replay (DGR), while training a solver ($S$), trains a generator ($G$) to produce past data when needed~\citep{shin2017continual,wu2018memory,ramapuram2020lifelong}, as illustrated in Fig.\ref{fig:1a}. However, the generative replay might be hindered in generating complex images~\citep{van2020brain}. 

Recently, CIL research has shifted from rebuilding the joint training environment to the knowledge distillation (KD)~\citep{hinton2015distilling}. It transfers the probabilistic knowledge between prediction layers from the old frozen network to a current trainable one. The solver network is usually split into two parts, a feature extractor and a classifier, corresponding to the extractor-aimed and prototype-aimed methods, respectively.

On the one hand,~\citet{rebuffi2017icarl} first introduce extractor-aimed methods that take the cross-entropy criterion to differentiate the learned classes and get a discriminative feature extractor. Instead of the cross-entropy criterion, various metric criteria have also been proposed.~\citet{yu2020semantic} enforce that the feature distance between relevant and irrelevant sample pairs is larger than a fixed margin.~\citet{cha2021co2l} constrain the contrastive relations between samples to be constant. In particular, Lucir establishes the cosine similarity between the extracted sample features and the parameters in the feature extractor~\citep{hou2019learning}. Benefiting from such similarity, Lucir not only obtains a discriminative feature extractor but could rectify the inter-class separations when confusion arises in CIL. However, all the above extractor-aimed methods require storing a few exemplars for each learned class~\citep{douillard2020podnet}.

For data privacy and security concerns, it is impossible to store any exemplars from the learned tasks~\citep{yu2020semantic, liu2020generative}. Therefore, on the other hand, prototype-aimed methods emerge. They store the feature mean (i.e., prototype), and keep the classifier behind the feature extractor to avoid the mismatch between the stored prototype and the feature extractor. For instance,~\citet{zhu2021prototype} incrementally expand the Gaussian distribution dominated by each stored prototype.~\citet{liu2020generative} argue that the prototypes cannot reasonably cover the actual feature distribution of classes. They hence introduce a generator to produce the learned features incrementally and suggest Generative Feature Replay (GFR) for CIL.

However, compared with the classifier that commonly consists of a few layers, the feature extractor is very large, and the conventional KD of the final predictions may not be enough for a suitable extractor. In contrast, richer knowledge embedded in the intermediate layers is lost during the CIL process. Besides, the purpose of the generative network, i.e., generating real instances, is different from that of the classifier, which aims to classify instances accurately. Similar to humans who cannot sketch (generate) an actual dollar bill but can readily spot (classify) a fake bill from the real ones, resolving this paradox between reality and discriminability~\citep{robert2016empty} may matter for CIL.

As such, we propose a novel CIL method in this paper based on a discriminative feature extractor and a well-designed prototype generation network. Specifically, we train the feature extractor incrementally in a layer-wise embedding distillation manner, which distills the embeddings of intermediate block outputs in the network to maintain more discriminative knowledge. Further, we devise a \textit{Ta}sk-oriented \textit{g}eneration (Tag) strategy, enabling a generative network to produce the proper features that preferably fit the needs of the classifier. As shown in Fig.\ref{fig:1b}, Tag trains the solver network $S$ first, then the generator network $G$ guided by the fixed $S$. In summary, our main contributions are as follows,
\begin{itemize}
    \item We propose an embedding distillation model for CIL to transfer the richer knowledge embedded in the network's intermediate blocks from the old frozen feature extractor to the current one. To our best knowledge, this is the first study to use embedding distillation with neither exemplars nor prototypes in CIL.
    \item We devise a task-oriented generation strategy to make the best of the trained classifier. We also empirically validate its merits of improving performances on CIL tasks.
    \item We propose a CIL method with \textit{e}mbedding distillation and \textit{Ta}sk-oriented \textit{g}eneration (eTag as shown in Fig.\ref{fig:1c}). It significantly outperforms SOTA methods on such large datasets as CIFAR-100 and ImageNet-sub.
\end{itemize}

\section{Related Work}

\subsection{Generative Replay in CIL}
\paragraph{Generative Sample Replay.} 
The core idea is to additionally train an incremental generator to produce samples of the learned task.~\citet{shin2017continual} first adopted the generative network to CIL and proposed the deep generative replay (DGR) method, as shown in Fig.\ref{fig:1a}. DGR, while training a solver ($S$) for inference, trained a generator ($G$) to produce past data when needed. Here, the network could be either Generative Adversarial Networks (GAN)~\citep{shin2017continual} or Variational Autoencoder (VAE)~\citep{ramapuram2020lifelong}. Instead of replaying the generated data, \citet{wu2018memory} implemented a memory replay mechanism by constraining the stored and current GANs with the same input-output pairs on the learned tasks. \citet{huang2022lifelong} developed a similar mechanism by extending VAE's intrinsic sample reconstruction property to knowledge reconstruction. BI-R~\citep{van2020brain} improved the training efficiency using VAE's symmetrical structure. As shown in Fig.\ref{fig:1b}, we improve the DGR structure to Tag and make the best of the solver to guide the training of the generator. Preliminary CIL experiments on the permutation MNIST~\citep{van2020brain} shown in Fig.\ref{fig:1d} verified its benefits compared with DGR.

\paragraph{Generative Feature Replay.}
To circumvent the limitations of DGR on generating the complex image, generative feature replay (GFR) focuses on estimating the feature distribution rather than the original image distribution. Note that the solver $S$ in GFR is split into a feature extractor $F$ and a classifier $C$, i.e., $S=F+C$. \citet{xiang2019incremental} employed a fixed feature extractor that is pre-trained on a large dataset and cannot be updated incrementally.~\citet{liu2020generative} distilled the final prediction from the stored feature extractor into the current trainable one, while updating the classifier with GFR. As shown in Fig.\ref{fig:1c}, we also explore CIL through generative feature replay, but we show eTag, a simple and efficient method of embedding distillation and task-oriented generation, performs significantly better than SOTA works.

\subsection{Embedding Distillation}
Embedding distillation~\citep{romero2014fitnets} extends the vanilla knowledge distillation~\citep{hinton2015distilling} from mimicking only the final prediction to the intermediate embeddings between the fixed network (i.e., teacher network) and the trainable one (i.e., student network). The main difficulty is determining what intermediate knowledge must be distilled and how to distill it. Various efforts have been made to address these two issues concurrently. \citet{ahn2019variational} reduced the knowledge gaps in intermediate layers between the student and teacher networks by maximizing the mutual information while~\citet{romero2014fitnets} aligned the intermediate feature maps using the mean square error. Similarly, \citet{heo2019knowledge} employed the mean square error but for aligning the activation boundaries derived from the hidden neurons. Recent works have introduced comparative information between samples to explore richer intermediate knowledge, e.g., \citep{peng2019correlation}, \citep{tian2019contrastive}, \citep{tung2019similarity}, and \citep{park2019relational}, etc. Beyond that, building additional self-supervised (SS) tasks to train the intermediate layer achieves better performance~\citep{yang2021hierarchical}. Although Podnet~\citep{douillard2020podnet} has verified that aligning the intermediate feature maps is beneficial for CIL, it stores exemplars, violating data privacy and security concerns. In this paper, we extend self-supervised embedding distillation to train the feature extractor, which is data-free without any exemplars.

\section{Proposed Method}
eTag owns two main modules: a feature generator $G$ and a solver $S$, where $S$ has a feature extractor $F$ and a classifier $C$. eTag incrementally trains $F$ and $C$ via embedding distillation (Sec. \ref{sec:3.1}) and task-oriented generative feature replay (Sec. \ref{sec:3.2}). Before going into depth, we first give the problem definition and framework overview.

\paragraph{Problem Definition.} We consider the CIL setting that a solver $S$ must sequentially learn $T$ tasks. Each task is expressed as a disjoint dataset, $\mathcal{D}_t=(\mathcal{X}_t, \mathcal{Y}_t)=\{(\boldsymbol{x}_t^i, y_t^i) \}_{i=1}^{n_t}, t=0, ..., T-1$, where $\mathcal{Y}_{t} \cap \mathcal{Y}_{t'} =\emptyset$ for all $t\neq t'$. $\boldsymbol{x}_t^i\in\mathcal{X}_t $ is the $i$-th image of class $y_t^i\in\mathcal{Y}_t=\{\mathcal{Y}_t^1, ..., \mathcal{Y}_t^{m_t} \}$, $n_t$ and $m_t$ are respectively the number of images and the number of classes. When training the new solver $S_t$ on task $t$, we could access the stored solver $S_{t-1}$ which is already trained on the task $(t-1)$ and the only available dataset $\mathcal{D}_t$. Afterward, the solver $S_t$ is expected to perform well in inference on all previously-learned tasks $t'\leq t$, where the task ID of each test sample is unknown.

\begin{figure*}[t]
    \centering
    \subfigure[Train $S_t=F_t+C_t$ with embedding distillation]{\label{fig:2a}\includegraphics[width=0.85\textwidth]{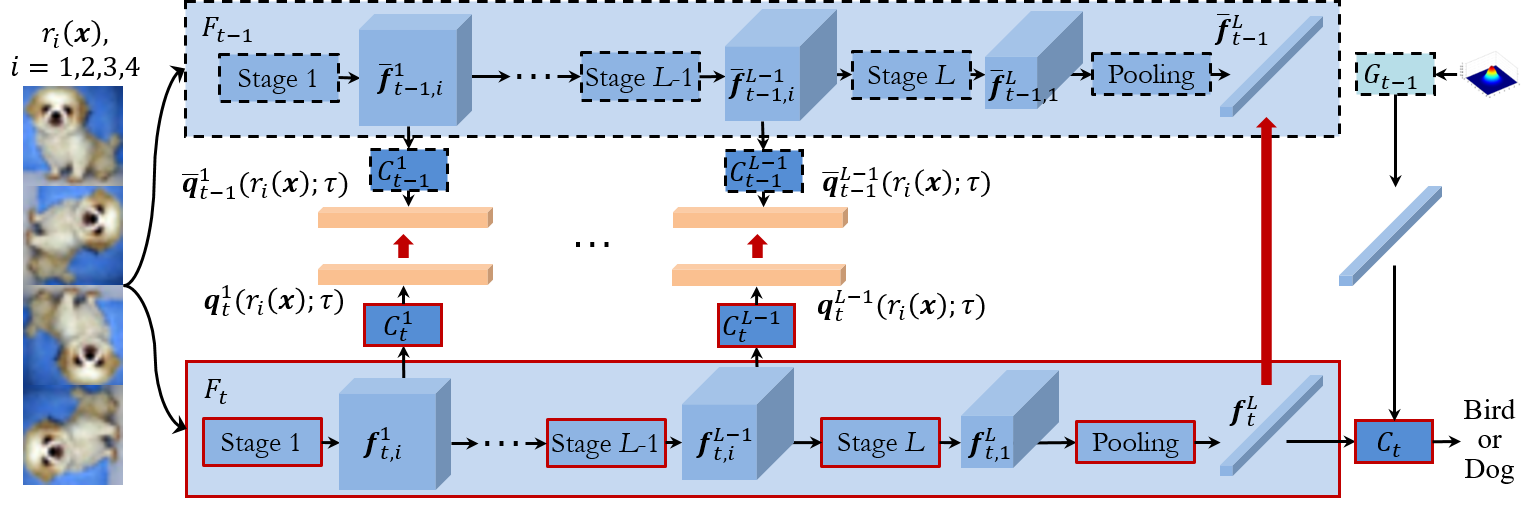}}  \\
    \subfigure[Train $G_t$ with task-oriented generation]{\label{fig:2b}\includegraphics[width=0.85\textwidth]{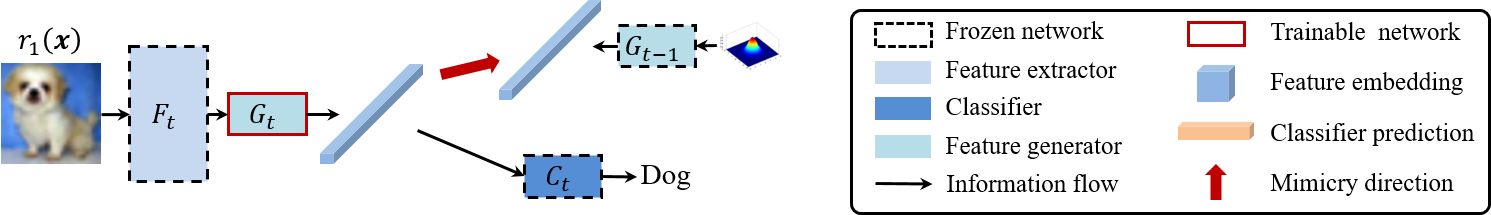}}
    \caption{Framework of the proposed eTag, including (a) incrementally training the solver $S_t=F_t+C_t$ with the help of the stored feature extractor $F_{t-1}$, generator $G_{t-1}$, and the only available dataset $\mathcal{D}_t$; (b) incrementally training the generator $G_t$ with the guidance from $C_t$.}
    \label{fig:2}
\end{figure*}

\paragraph{Framework Overview.} As illustrated in Fig.\ref{fig:2}, we use ResNet-18~\citep{he2016deep}, a modern staged convolutional neural network, as our solver, and VAE as our feature generator\footnote{It is straightforward to generalize the solver to other modern convolutional neural networks, such as VGG~\citep{simonyan2014very}, MobileNet~\citep{sandler2018mobilenetv2}, etc. Compared with GAN, VAE has a more stable training mechanism~\citep{huang2022lifelong,ramapuram2020lifelong}.}. Assuming the feature extractor contains $L$ stages, we append an auxiliary classifier behind each stage except the final one, yielding $(L-1)$ auxiliary classifiers $\{C^l(\cdot)\}_{l= 1}^{L-1}$. $C^l(\cdot)$ is composed of stage-wise convolutional blocks, a global average pooling layer, and a fully connected layer~\citep{yang2021hierarchical}. Note that these auxiliary classifiers are discarded during inference, since they are only engaged to assist the staged embedding distillation. The final classifier $C$ consists of a linear connected layer $\boldsymbol{w}$ and a non-linear activation layer $\sigma$. For the feature $\boldsymbol{f}$ of a given sample, $C(\boldsymbol{f})=\sigma(\boldsymbol{w}\cdot \boldsymbol{f})\in\mathbb{R}^{1\times m_t}$. We train the feature extractor $F_t$ and classifier $C_t$ on task $t$ using embedding distillation and feature generative replay. After that, we freeze $F_t$ and $C_t$, and train the feature generator $G_t$ with guidance from the classifier $C_t$, i.e., task-oriented generation, as shown in Fig.\ref{fig:2b}. 

\subsection{Embedding Distillation} \label{sec:3.1}
Embedding distillation includes two operations on the embedded knowledge: construction and distillation. Specifically, embedding knowledge construction allows the feature extractor to learn the current task well, while embedding knowledge distillation enables the feature extractor to remember the learned tasks.

\paragraph{Embedding Knowledge Construction.} 
For the $L$-staged feature extractor, we separate the embedded knowledge into a final output feature, $\boldsymbol{f}^L$, and $(L-1)$ embeddings, $\{\boldsymbol{f}^{l}\}_{l=1}^{L-1}$.

On the one hand, in order to enable the feature extractor, $F_{t-1}$, to extract a discriminative final feature, $\boldsymbol{f}^L_{t-1}$, on task $(t-1)$, we train the solver, $S_{t-1}=F_{t-1}+C_{t-1}$, with the conventional cross-entropy loss,
\begin{align} \label{eq:1}
    &\mathcal{L}_{CE}^{final}(F_{t-1}, C_{t-1}) \nonumber \\
    =&\mathbb{E}_{\boldsymbol{x} \in \mathcal{X}_{t-1} ,y \in \mathcal{Y}_{t-1}}\mathcal{L}_{CE}\left( C_{t-1} \circ F_{t-1} (\boldsymbol{x} ), y\right).
\end{align}

On the other hand, to obtain $(L-1)$ embeddings, $\{\boldsymbol{f}_{t-1}^{l}\}_{l=1}^{L-1}$, from $F_{t-1}$, we train the auxiliary classifiers $\{C_{t-1}^l(\cdot)\}_{l=1}^{L-1}$ on the constructed additional self-supervised task. Specifically, given an image $\boldsymbol{x}\in\mathcal{X}_{t-1}$ and its label $y\in\mathcal{Y}_{t-1}$, we employ $4$ rotations (i.e., $0\degree, 90\degree, 180\degree, 270\degree$) of $\boldsymbol{x}$, $r_i(\boldsymbol{x}), i=1,2,3,4$, and the augmentation labels of $y$, i.e., $y_{r_i}$, as our additional self-supervised task, where $r_1(\boldsymbol{x})=\boldsymbol{x}$,  $y_{r_1}=y$.
The prediction of $r_i(\boldsymbol{x})$ on the $l$-th auxiliary classifiers $C^l_{t-1}$ is $\boldsymbol{q}_{t-1}^{l}\left( r_{i}( \boldsymbol{x}); \tau\right)=\sigma\left(C_t^l(\boldsymbol{f}_{t-1,i}^l)\right)\in\mathbb{R}^{1\times 4\cdot m_t}$. $\tau$ is a hyper-parameter to scale the smoothness of distribution~\citep{hinton2015distilling}; $\boldsymbol{f}_{t-1,i}^l$ is the $l$-th intermediate embedding of the $i$-th rotated image $r_i(\boldsymbol{x})$. We train $(L-1)$ auxiliary classifiers with $\mathcal{L}_{CE}^{inter}$ that fits the rotated $r_i(\boldsymbol{x})$ to the augmentation label $y_{r_i}$,
\begin{align} \label{eq:2}
    &\mathcal{L}_{CE}^{inter}(F_{t-1}, C_{t-1}^l) \nonumber \\
    =&\frac{1}{4}\sum _{i=1}^{4}\sum_{l=1}^{L-1}\mathcal{L}_{CE}\left( \sigma \left( C_{t-1}^{l} (\boldsymbol{f}_{t-1,i}^{l} )\right) ,\ y_{r_{i}}\right).
\end{align}
Eventually, $\boldsymbol{f}^L_{t-1}$ and $\{\boldsymbol{f}_{t-1}^{l}\}_{l=1}^{L-1}$ could embed more knowledge from data $\mathcal{D}_{t-1}$ compared with the vanilla training with $\boldsymbol{f}^L_{t-1}$ only.

\paragraph{Embedding Knowledge Distillation.}
When a new task $t$ arrives, we employ the embedding knowledge distillation to retain the knowledge of the feature extractor learned before.

Firstly, we distill the final feature distribution from $F_{t-1}$ to $F_t$ using $\mathcal{L}_{2}$ loss~\citep{liu2020generative}, 
\begin{align} \label{eq:3}
    \mathcal{L}_{2}^{final}(F_t)
    &=\mathbb{E}_{\boldsymbol{x} \in \mathcal{X}_{t}}[\Vert F_{t}(\boldsymbol{x}) -F_{t-1}(\boldsymbol{x})\Vert _{2}],
\end{align}
where $F_{t-1}(\boldsymbol{x})=\boldsymbol{\bar{f}}_{t-1}^L$ is the final feature extracted from $F_{t-1}$ on task $t$. Note that we use a bar, $\bar{\square}$, to denote the outputs of the $(t-1)$-th model on the data of the current task.

Secondly, we distill $(L-1)$ intermediate embeddings from $F_{t-1}$ to $F_{t}$ with a stage-wise Kullback–Leibler divergence loss,
\begin{small}
\begin{align} \label{eq:4}
    &\mathcal{L}_{KL}^{inter}(F_t, C_t^{l}) \\
    =&\frac{1}{4}\sum_{i=1}^{4}\sum _{l=1}^{L-1} \tau ^{2}\mathcal{L}_{KL}\left( \sigma \left( C_{t-1}^{l} (\boldsymbol{\bar{f}}_{t-1,i}^{l} )/\tau\right)\Big\Vert\sigma\left( C_{t}^{l} (\boldsymbol{f}_{t,i}^{l} )/\tau \right)\right),\nonumber
\end{align}
\end{small}
where $\tau=3$ in $\mathcal{L}_{KL}^{inter}$, and $\tau^2$ is utilized to keep the gradient contributions unchanged~\citep{hinton2015distilling}.

In addition, we train the solver $S_t=F_t+C_t$ with $\mathcal{L}_{CE}$ on $\mathcal{D}_{t}$ to render $F_t$ plastic to learn the incremental task $t$. Simultaneously, by fitting the generated feature $\boldsymbol{\hat{f}}^L_{:t-1}$ to its corresponding label $y\in\mathcal{Y}_{:t-1}$, $\mathcal{L}_{CE}$ loss enables the classifier $C_t$ to stably retain the learned tasks,
\begin{align}
  &\mathcal{L}_{CE}^{new}( F_{t} ,C_{t}) = \mathcal{L}_{CE}^{final}( F_{t} ,C_{t}) \\
  &+\lambda_{CE}\cdot\mathbb{E}_{\boldsymbol{\hat{f}}_{:t-1}^{L} ,\ y \in \mathcal{Y}_{:t-1}}\mathcal{L}_{CE}\left( \sigma \left(\boldsymbol{w}_{:t-1} \cdot \boldsymbol{\hat{f}}_{:t-1}^{L}\right), y\right),  \nonumber
\end{align}
where $\hat{\square}$ indicates the generated output, $\mathcal{Y}_{:t-1}$ represents all labels of $(t-1)$ learned tasks, and $\lambda_{CE}$ is automatically set to $m_{:t-1}/{m_t}$, i.e., the ratio of the number of classes in all learned tasks to that in the current task.

\subsection{Task-Oriented Generation} \label{sec:3.2}
We incrementally train a feature generator $G_{t}$ to estimate the conditional distribution of features, $p(\boldsymbol{\hat{f}}^L_{:t}\mid y_{:t})=G_t(y_{:t}, \boldsymbol{z})$, preventing the classifier $C_t$ from being forgotten. Here, $y_{:t}$ is a sample's label from the tasks learned so far. The Gaussian noise vector $\boldsymbol{z}$ is used to stimulate $G_{t}$ to produce old features.

For the current task, we specifically propose a task-oriented generation strategy, which allows $G_{t}$ to produce features that better fit the needs of the classifier, $C_t$. We adopt the conditional VAE network to realize this purpose with the following modified VAE loss,
\begin{align} \label{eq:6}
  & \mathcal{L}_{VAE}^{new}( G_{t}) = \mathcal{L}_{KL}\left( p_{\phi _{t}}\left(\boldsymbol{z}\mid\boldsymbol{f}_{t}^{L} ,y\right) \big\Vert p(\boldsymbol{z})\right) \\
  &-\mathbb{E}_{p_{\phi _{t}}\left(\boldsymbol{z}\vert\boldsymbol{f}_{t}^{L} ,y\right), \boldsymbol{x} \in \mathcal{X}_{t} , y\in \mathcal{Y}_{t}}\left[\log p_{\theta_{t}}\left(\boldsymbol{\hat{f}}_{t}^{L} \big\vert y,\boldsymbol{z}\right)\right] \nonumber \\
  &+\mathbb{E}_{p_{\theta _{t}}\left(\boldsymbol{\hat{f}}_{t}^{L} \mid y,\boldsymbol{z}\right), y\in \mathcal{Y}_{t}, \boldsymbol{z} \sim p(\boldsymbol{z})}\mathcal{L}_{CE}\left( \sigma \left(\boldsymbol{w}_t \cdot \boldsymbol{\hat{f}}_{t}^{L}\right), y\right),\nonumber
\end{align}
where $\phi_t$ and $\theta_t$ indicate, respectively, the parameters of encoder and decoder in $G_t$; and $p(\boldsymbol{z})=\mathcal{N}(0,1)$, i.e., the standard Gaussian distribution. By minimizing the first two terms, the encoder promotes $\boldsymbol{f}_{t}^{L}$ and $y$ to be the prior Gaussian distribution, while the decoder reconstructs $y$ and $\boldsymbol{z}$ back to the original feature $\boldsymbol{f}_{t}^{L}$. The last term encourages the generator $G_t$ to produce features such that the classified label of $C_t$ is correct, improving the coherence of the feature from $p_{\theta_t}$ with the label $y$.

To prevent forgetting the learned $(t-1)$ tasks, we use the knowledge reconstruction strategy~\citep{huang2022lifelong} that enables $G_t$ to reconstruct the historical knowledge retained in $G_{t-1}$. Specifically, the following knowledge reconstruction loss is appended,
\begin{align}
   \mathcal{L}_{VAE}^{old}( G_{t}) = -&\mathbb{E}_{p_{\theta _{t-1}}\left(\boldsymbol{\hat{f}}_{:t-1}^{L} \mid y,\boldsymbol{z}\right),\ y\sim U\{\mathcal{Y}_{:t-1}\},\ \boldsymbol{z} \sim p(\boldsymbol{z})} \nonumber \\
   & \left[\log p_{\theta _{t}}\left(\boldsymbol{\hat{f}}_{:t-1}^{L}\mid y,\boldsymbol{z}\right)\right],
\end{align}
where $U\{\cdot\}$ is the discrete uniform distribution; and the label $y$ is uniformly sampled from the labels of learned tasks.
\begin{algorithm}[!t]
\DontPrintSemicolon
\KwIn{A sequence of $T$ task: $\mathcal{D}_0, ..., \mathcal{D}_{T-1}$, where $\mathcal{D}_t=(\mathcal{X}_t, \mathcal{Y}_t)$}
\KwOut{Incremental solver $S_T=(F_T, C_T)$}
\SetKwBlock{Begin}{function}{end function}
\Begin($\text{eTag}{(} \mathcal{D}_1, ..., \mathcal{D}_T {)}$)
{
    Train $F_0$, $C_0$, and $C_0^l$ on $\mathcal{D}_0$ using Eq.\eqref{eq:8}  \;
    Train $G_0$ on $\mathcal{D}_0$ using Eq.\eqref{eq:6} \;
    \textit{Evaluate $F_0$ and $C_0$ on the initial task $0$} \; 
    \For{$t = 1,...,T-1$}
    {
        Train $F_t$, $C_t$, and $C_t^l$ on $\mathcal{D}_t$ using Eq.\eqref{eq:10} \;
        Train $G_t$ on $\mathcal{D}_t$ using Eq.\eqref{eq:11} \;
        \textit{Evaluate $F_t$ and $C_t$ on the learned task $1:t$} \;
    }
  \Return{$F_T$ and $C_T$}
}
\caption{CIL with eTag}\label{alg:1}
\end{algorithm}
\begin{table*}[!t]
  \caption{Overall average incremental accuracy ($A$) and forgetting measure ($F$) on CIFAR-100 and ImageNet-sub after the model incrementally learns \# tasks. \colorbox{Gray}{$e$XX} stands for the modified method that introduces our proposed idea of \textit{embedding distillation} in \colorbox{Gray}{XX}. \textbf{Bold} and \underline{underline} indicate the optimal and sub-optimal results, respectively. We only report the {\tiny($\pm$std)} results of $25$ tasks due to the space limitation.}
  \label{tab:1}
  \centering
  \resizebox{0.998\linewidth}{!}{
    \begin{tabular}{c|cc|cc|cc|cc|cc|cc}
    \toprule
    \multirow{2}*{Dataset} & \multicolumn{6}{c|}{CIFAR-100} & \multicolumn{6}{c}{ImageNet-sub} \\
    \cmidrule(lr){2-7} \cmidrule(lr){8-13} 
      & \multicolumn{2}{c}{5 tasks} & \multicolumn{2}{c}{10 tasks} & \multicolumn{2}{c|}{25 tasks} & \multicolumn{2}{c}{5 tasks} & \multicolumn{2}{c}{10 tasks} & \multicolumn{2}{c}{25 tasks} \\
    \cmidrule(lr){1-1} \cmidrule(lr){2-3} \cmidrule(lr){4-5} \cmidrule(lr){6-7} \cmidrule(lr){8-9} \cmidrule(lr){10-11} \cmidrule(lr){12-13}
    Metric  & $A (\uparrow)$ & $F (\downarrow)$ & $A (\uparrow)$ & $F (\downarrow)$ & $A (\uparrow)$ & $F (\downarrow)$
            & $A (\uparrow)$ & $F (\downarrow)$ & $A (\uparrow)$ & $F (\downarrow)$ & $A (\uparrow)$ & $F (\downarrow)$ \\
    \midrule
    Joint	& 75.21 & 3.79 & 75.26 & 4.09 & 75.11{\tiny$(\pm1.13)$} & 6.60{\tiny$(\pm1.29)$} & 78.92 & 3.84 & 79.53 & 6.27 & 80.12{\tiny$(\pm0.25)$} & 8.74{\tiny$(\pm0.20)$} \\
    Fine    & 22.59 & 65.49 & 12.20 & 72.59 & 4.66{\tiny$(\pm0.06)$} & 76.64{\tiny$(\pm0.73)$} & 19.53 & 75.20 & 9.58 & 79.40 & 4.60{\tiny$(\pm0.03)$} & 80.13{\tiny$(\pm0.62)$} \\
    \midrule
    EWC		& 31.66 & 56.10 & 21.41 & 62.51 & 7.93{\tiny$(\pm0.56)$} & 73.18{\tiny$(\pm0.65)$}	& 30.53 & 59.18 & 21.00 & 66.84 & 10.87{\tiny$(\pm0.12)$} & 73.62{\tiny$(\pm0.51)$} \\
    \rowcolor{Gray}
    $e$EWC & 36.39 & 53.67 & 23.21 & 60.02 & 7.86{\tiny$(\pm0.52)$} &	75.67{\tiny$(\pm1.63)$} & 35.10 & 59.52 & 21.50 & 69.41 & 10.68{\tiny$(\pm0.41)$} & 10.68{\tiny$(\pm0.41)$}  \\
    MAS     & 25.95 & 63.78 & 17.43 & 67.93 & 7.87{\tiny$(\pm1.69)$} & 74.45{\tiny$(\pm1.06)$}	& 29.20 & 63.39 & 17.93 & 70.66 & 8.13{\tiny$(\pm0.81)$} & 76.25{\tiny$(\pm0.38)$} \\
    \rowcolor{Gray}
    $e$MAS & 31.64 & 59.17 & 20.69 & 62.43 & 7.13{\tiny$(\pm0.97)$} & 76.89{\tiny$(\pm2.11)$} & 32.53 & 62.43 & 19.76 & 71.11 & 8.56{\tiny$(\pm0.66)$} & 78.51{\tiny$(\pm1.17)$} \\
    LwF & 52.98 & 31.18 & 42.27 & 39.51 & 16.92{\tiny$(\pm1.21)$} &	63.89{\tiny$(\pm1.32)$} & 54.59 & 33.11 & 44.60 & 40.76 & 26.41{\tiny$(\pm0.47)$} & 57.61{\tiny$(\pm0.42)$} \\
    \rowcolor{Gray}
    $e$LwF & 52.16 & 35.77 & 40.58 & 44.93 & 19.45{\tiny$(\pm0.84)$} & 64.88{\tiny$(\pm1.83)$} & 47.43 & 45.35 & 44.26 & 44.67 & 23.37{\tiny$(\pm1.18)$} & 63.64{\tiny$(\pm1.41)$} \\
    LwM & 56.95 & 26.60 & 44.85 & 38.08 & 25.40{\tiny$(\pm1.36)$} & 55.47{\tiny$(\pm1.32)$} & 52.90 & 34.92 & 42.86 & 42.44 & 26.41{\tiny$(\pm0.39)$} & 57.28{\tiny$(\pm0.98)$} \\
    \rowcolor{Gray}
    $e$LwM & 57.94 & 28.83 & 45.86 & 40.34 & 24.71{\tiny$(\pm0.52)$} & 59.39{\tiny$(\pm1.24)$} & 50.60 & 41.18 & 41.74 & 47.39 & 25.03{\tiny$(\pm0.16)$} & 62.08{\tiny$(\pm0.15)$} \\
    IL2M    & 53.83 & 30.78 & 53.19 & 29.00 & 46.45{\tiny$(\pm0.39)$} & 33.83{\tiny$(\pm2.06)$} & 55.89 & 31.12 & 56.63 & 26.46 & 52.85{\tiny$(\pm0.59)$} & 29.71{\tiny$(\pm0.94)$} \\
    \rowcolor{Gray}
    $e$IL2M & 58.63 & 27.88 & 55.80 & 31.35 & 47.66{\tiny$(\pm1.84)$} & 35.58{\tiny$(\pm2.87)$} & 54.80 & 36.39 & 53.16 & 35.21 & 50.17{\tiny$(\pm1.59)$} & 35.50{\tiny$(\pm1.75)$} \\
    Lucir   & 56.95 & \underline{18.74} & 52.87 & \underline{21.71} & 49.35{\tiny$(\pm2.66)$} & 24.83{\tiny$(\pm1.83)$}	& 68.23 & \textbf{13.69} & 66.90 & \textbf{14.83} & 64.83{\tiny$(\pm0.29)$} & 19.90{\tiny$(\pm0.82)$} \\
    \rowcolor{Gray}
    $e$Lucir & 65.06 & 20.77 & 60.60 & 26.93 & 51.69{\tiny$(\pm2.16)$} & 32.00{\tiny$(\pm3.45)$} & 70.49 & 18.27 & 67.61 & 19.94 & 61.21{\tiny$(\pm3.83)$} & 25.05{\tiny$(\pm3.16)$} \\
    GFR     & 62.74 & 19.69 & 59.85 & 21.74 & \underline{56.76}{\tiny$(\pm1.62)$} & \underline{23.21}{\tiny$(\pm1.62)$} & 69.91 & 18.24 & 67.30 & 19.52 & 63.70{\tiny$(\pm0.61)$} & 22.23{\tiny$(\pm0.48)$} \\
    \rowcolor{Gray}
    $e$GFR & \underline{66.76} & 19.03 & \underline{64.62} & 23.95 & 55.20{\tiny$(\pm0.94)$} & 28.53{\tiny$(\pm1.58)$} & \underline{76.14} & 14.76 & \underline{74.24} & 15.28 & \underline{70.33}{\tiny$(\pm0.17)$} & \underline{18.37}{\tiny$(\pm0.22)$} \\
    eTag & \textbf{67.99} & \textbf{16.89} & \textbf{65.50} & \textbf{18.01} & \textbf{61.63}{\tiny$(\pm0.79)$} & \textbf{21.95}{\tiny$(\pm0.74)$} & \textbf{76.79} & \underline{14.50} & \textbf{75.17} & \underline{15.04} & \textbf{71.77}{\tiny$(\pm0.12)$} & \textbf{17.86}{\tiny$(\pm0.27)$} \\
    \bottomrule
    \end{tabular}
  }
\end{table*}

\subsection{Integrated Objective}
Our method addresses the forgetting problem in CIL.
When learning the initial task, there is no previous knowledge to retain. Hence we train the initial generator with $\mathcal{L}_{VAE}^{new}( G_{0})$, and the solver with,
\begin{align} \label{eq:8}
  \mathcal{L}\left( F_{0}, C_{0}, C_{0}^{l}\right) = \mathcal{L}_{CE}^{final}( F_{0}, C_{0}) +\mathcal{L}_{CE}^{inter}\left( F_{0} ,C_{0}^{l}\right).
\end{align}
When learning the task $t\geq 1$, we train the solver and generator incrementally with the losses as below,
\begin{align} \label{eq:10}
    &\mathcal{L}\left( F_{t} ,\ C_{t} ,\ C_{t}^{l}\right)\\ 
    =&\mathcal{L}_{CE}^{new}( F_{t} ,\ C_{t}) +\lambda \cdot \left(\mathcal{L}_{2}^{final}( F_{t}) +\mathcal{L}_{KL}^{inter}\left( F_{t} ,C_{t}^{l}\right)\right), \notag
\end{align}
\begin{align} \label{eq:11}
    \mathcal{L}( G_{t}) =\mathcal{L}_{VAE}^{new}( G_{t}) +\lambda_{VAE} \cdot \mathcal{L}_{VAE}^{old}( G_{t}),
\end{align}
where parameters $\lambda$ and $\lambda_{VAE}$ are also automatically set to the same value as $\lambda_{CE}$, i.e., $\lambda=\lambda_{VAE}=m_{:t-1}/{m_t}$. We summarize the whole training process in Algorithm~\ref{alg:1}.
\begin{figure*}[!t]
	\centering
	\subfigure[5 incremental tasks]{\label{fig:detacca}\includegraphics[height=4.4cm]{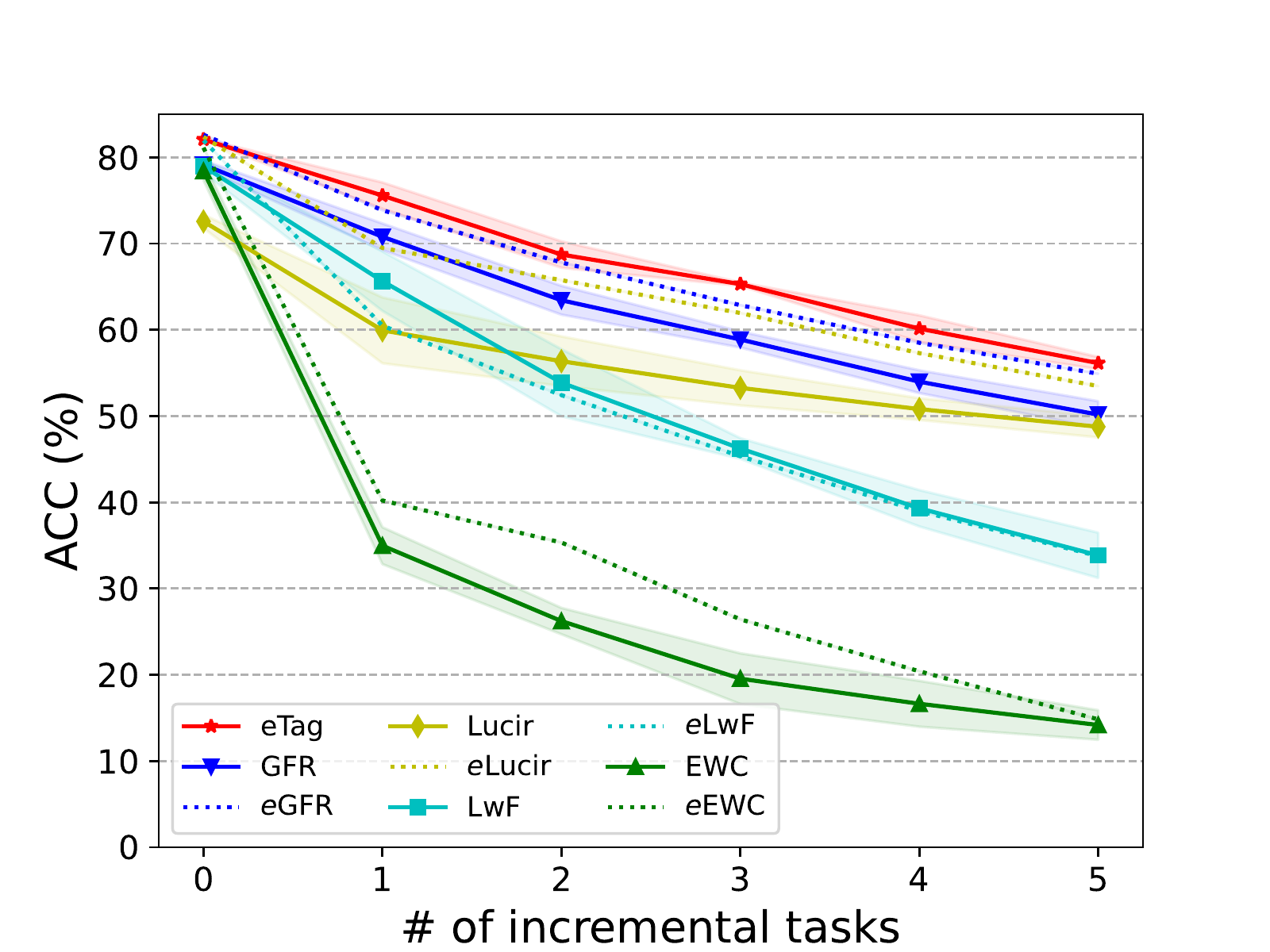}}
	\hspace*{\fill}
	\subfigure[10 incremental tasks]{\label{fig:detaccb}\includegraphics[height=4.4cm]{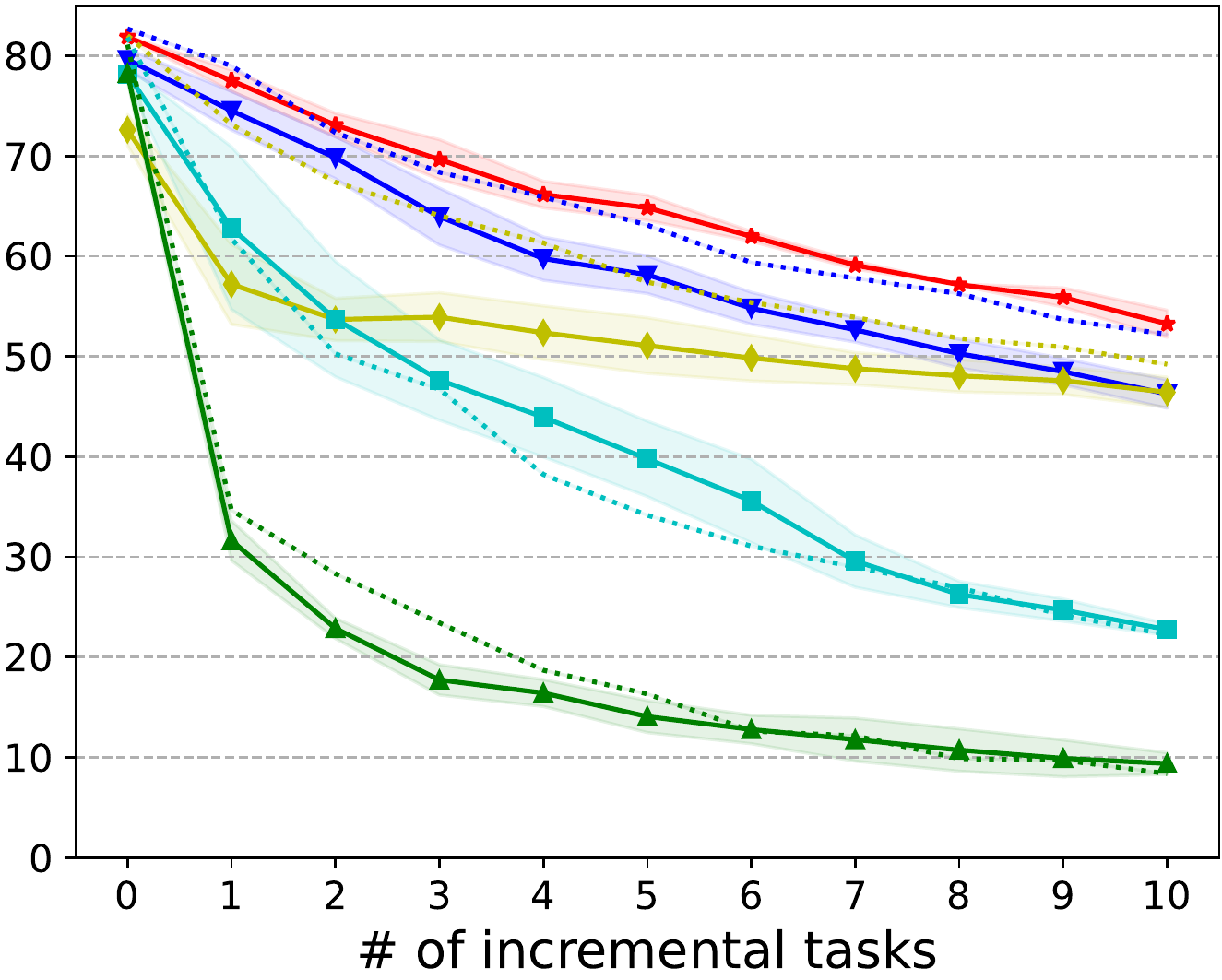}}
	\hspace*{\fill}
	\subfigure[25 incremental tasks]{\label{fig:detaccc}\includegraphics[height=4.4cm]{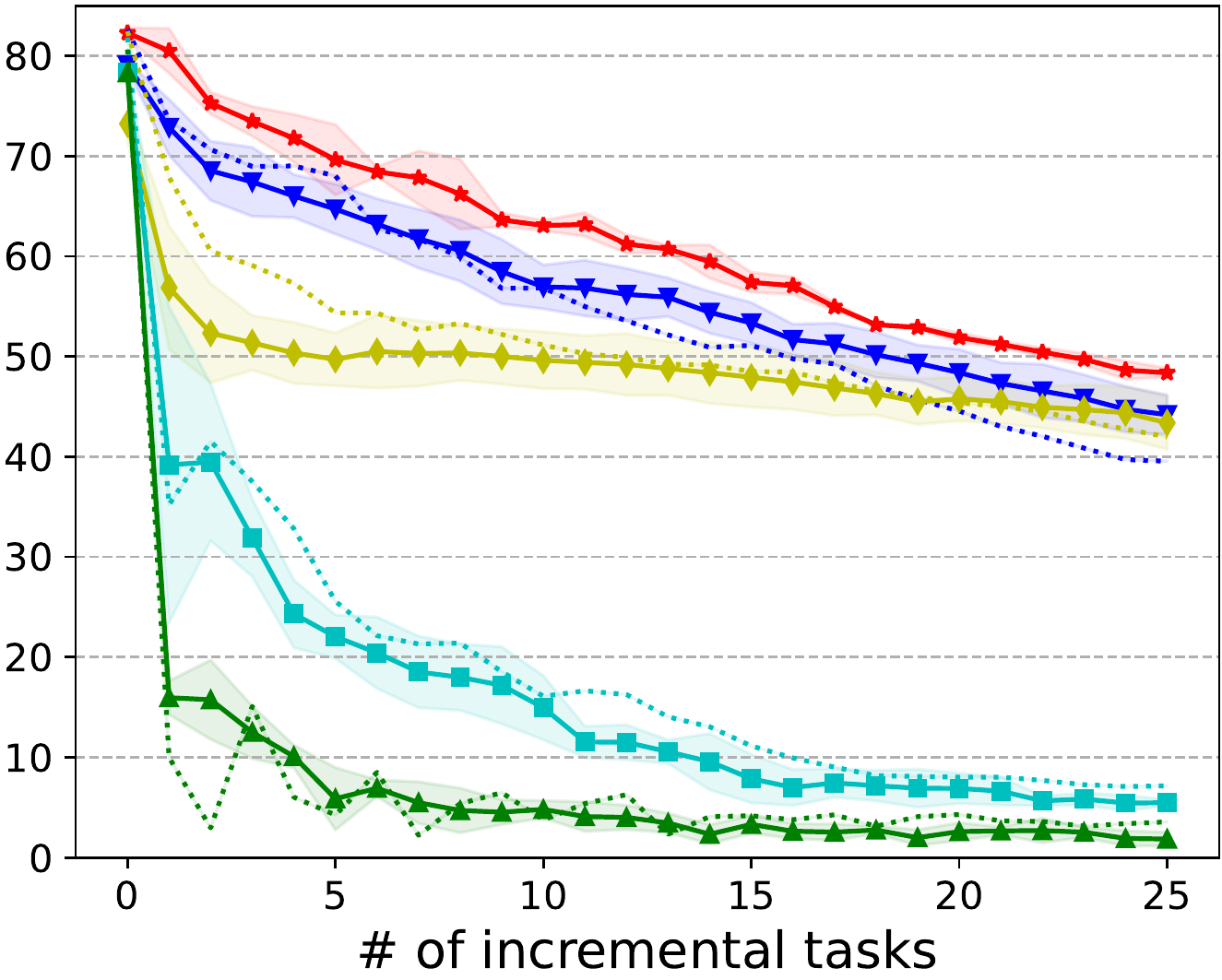}}
	\caption{Detailed ACCs on CIFAR-100 with $5$, $10$, and $25$ class-incremental tasks. The results on ImageNet-sub are given in Appendix~\ref{app:2}.}
	\label{fig:detacc}
\end{figure*}

\section{Experimental Results}
\paragraph{Datasets.}
We carry out the experiments on CIFAR-100~\citep{krizhevsky2009learning} and ImageNet-sub~\citep{deng2009imagenet}. CIFAR-100 contains $60,000$ images for $100$ classes, where each class has $500$ images for training and $100$ for testing. Each image is $32\times32$ pixels in size. We pad CIFAR-100 images with $4$ pixels, randomly sample $32\times32$ crops during training, and center crops for testing. ImageNet-sub contains $100$ classes randomly selected from ImageNet. Each class has more than $1,000$ images for training and $500$ for testing. Each image is resized to $256\times256$ before randomly sampling $224\times224$ crops for training. For testing, the original center crop with $224\times224$ shapes is used.

\paragraph{Evaluation Metrics.}
We employ \textit{average incremental accuracy} ($A$), \textit{average forgetting measure} ($F$),  and classification \textit{accuracies} (ACC) for evaluation~\citep{rebuffi2017icarl,chaudhry2018riemannian,hou2019learning}. $A$ (higher indicates better performance) and $F$ (lower indicates less forgetting) are calculated averagely over all learned tasks, while the detailed ACC is calculated on per learned task~\citep{yu2020semantic}.

\paragraph{Compared Methods.} We compare two reference methods, fine-tuning (Fine) and joint training (Joint)~\citep{chen2018lifelong}, and several SOTA methods on CIFAR-100 and ImageNet-sub. They include two parameter isolation methods, LwF~\citep{li2017learning} and LwM~\citep{dhar2019learning}, two regularization-based methods, EWC~\citep{kirkpatrick2017overcoming} and MAS~\citep{aljundi2018memory}, two extractor-aimed methods, IL2M~\citep{belouadah2019il2m} and Lucir~\citep{hou2019learning}, and the prototype-aimed method, GFR~\citep{liu2020generative}. Besides, we directly introduce the embedding distillation into each SOTA method to obtain a modified $e$XX method, where XX is LwF, LwM, MAS, IL2M, Lucir, or GFR.

Fine updates the network on the newly arrived tasks, while Joint assumes data of all previous tasks are available in each training task. EWC, MAS, LwF, LwM, and GFR are trained without exemplars, whereas IL2M and Lucir store $20$ exemplars for each learned class. We train a multi-head network for EWC, MAS, LwF, and LwM, and take the inference result with the highest probability across all heads. For GFR, we use the same VAE generator as ours to produce features. Other parameters of the comparisons are consistent with that in~\citep{masana2020class}. All experiments are carried out with $3$ different seeds. Please see Appendix~\ref{app:a1} for detailed implementations.

\subsection{CIL Evaluations} \label{subsec:cileva}
We begin with a model trained on half of the classes in a dataset, and the remaining classes arrive in $5$, $10$, and $25$ incremental tasks, similar to~\citep{hou2019learning,liu2020generative}.

\paragraph{Overall Results.} 
Tab.\ref{tab:1} shows that eTag outperforms all comparisons in incremental accuracy by a large margin, which on total $100$ classes improves by more than $5\%$ for CIFAR-100 and $7\%$ for ImageNet-sub compared with the sub-optimal results of GFR. Correspondingly, eTag suffers less from forgetting than others, indicating its competitive performance. The main reasons are that we construct extra self-supervised tasks on the current data to distill the embedding knowledge and take Tag to produce specific features for incrementally training the classifier. Interestingly, directly introducing the embedding distillation into the SOTA methods did improve their incremental accuracy, but did not reduce their degree of forgetting measures. For example, on $10$ tasks of ImageNet-sub, {$e$Lucir} improves Lucir with $0.71$ incremental accuracy while suffering more serious forgetting from $14.83\%$ to $19.94\%$. It mainly attributes to the fact that Tag generates stable features, enabling eTag to enjoy better accuracy and less forgetting.

\paragraph{Detailed ACCs.}
We select EWC, LwM, Lucir, and GFR for clarity, since they yield better incremental accuracy in each category of comparisons. Fig.\ref{fig:detacc} shows that EWC and LwM perform unsatisfactorily, especially with more incremental tasks. Lucir gets reasonable ACCs with stored exemplars, but the cosine criterion degrades its performance on the first task. In contrast, GFR uses the cross-entropy criterion to extract features and obtains comparable results to Lucir, even without any exemplar. Besides, the results verify that directly plugging in the embedding distillation in these comparisons is not always beneficial for each incremental task. Such as for $e$GFR on $25$ incremental tasks, the ACCs drop a lot compared with the original GFR. By unifying Tag and embedding distillation, eTag stably attains exceeded performances in all settings with SOTA methods by a large margin.

\begin{figure*}[!tb]
  \centering
  \resizebox{0.998\linewidth}{!}{
  \subfigure[Task 0]{\label{fig:conmata}\includegraphics[height=3.5cm]{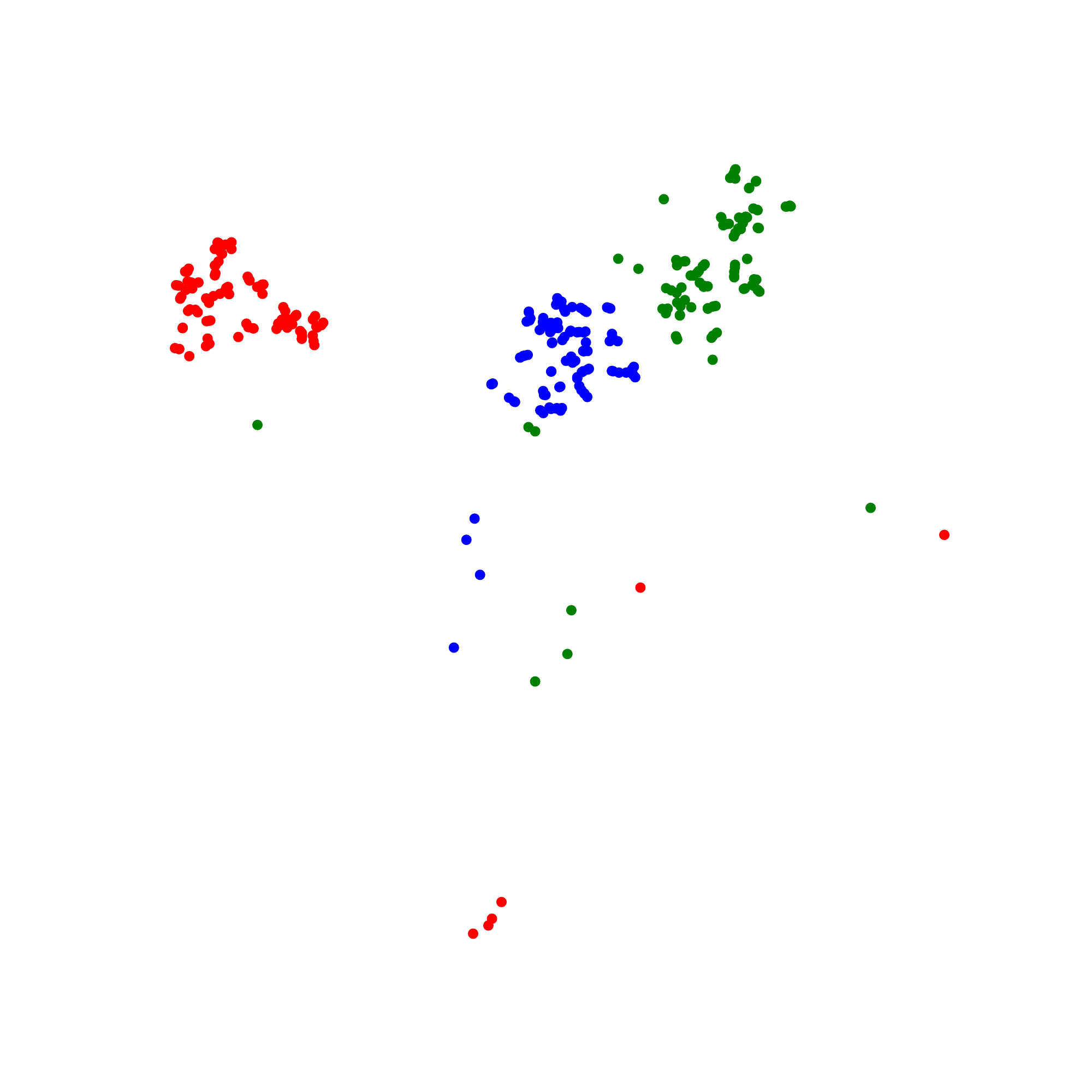}}
  \tikz{\draw[densely dotted](0, 3.5) -- (0,0);}
  \subfigure[Task 1]{\label{fig:conmatb}\includegraphics[height=3.5cm]{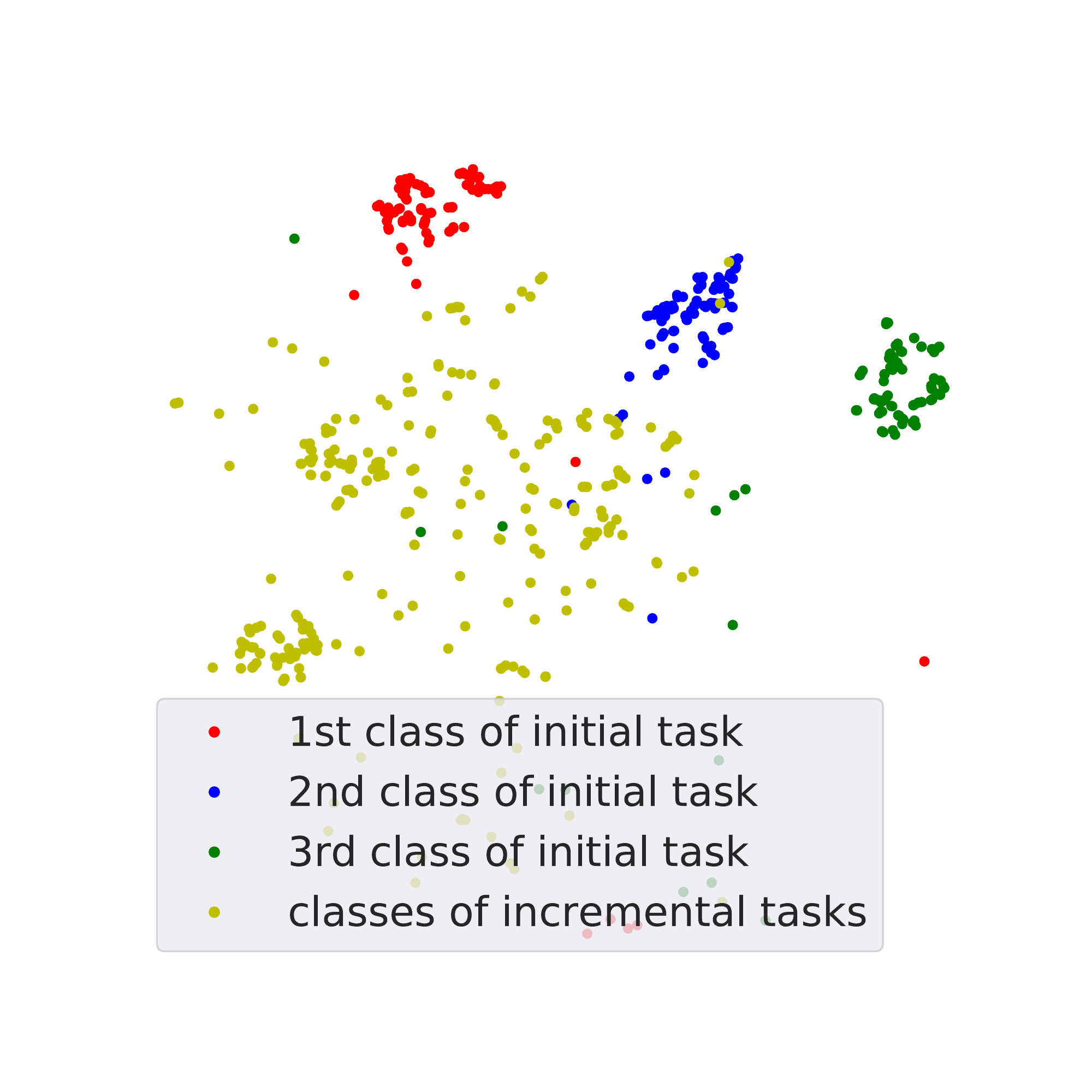}}
  \tikz{\draw[densely dotted](0, 3.5) -- (0,0);}
  \subfigure[Task 2]{\label{fig:conmatc}\includegraphics[height=3.5cm]{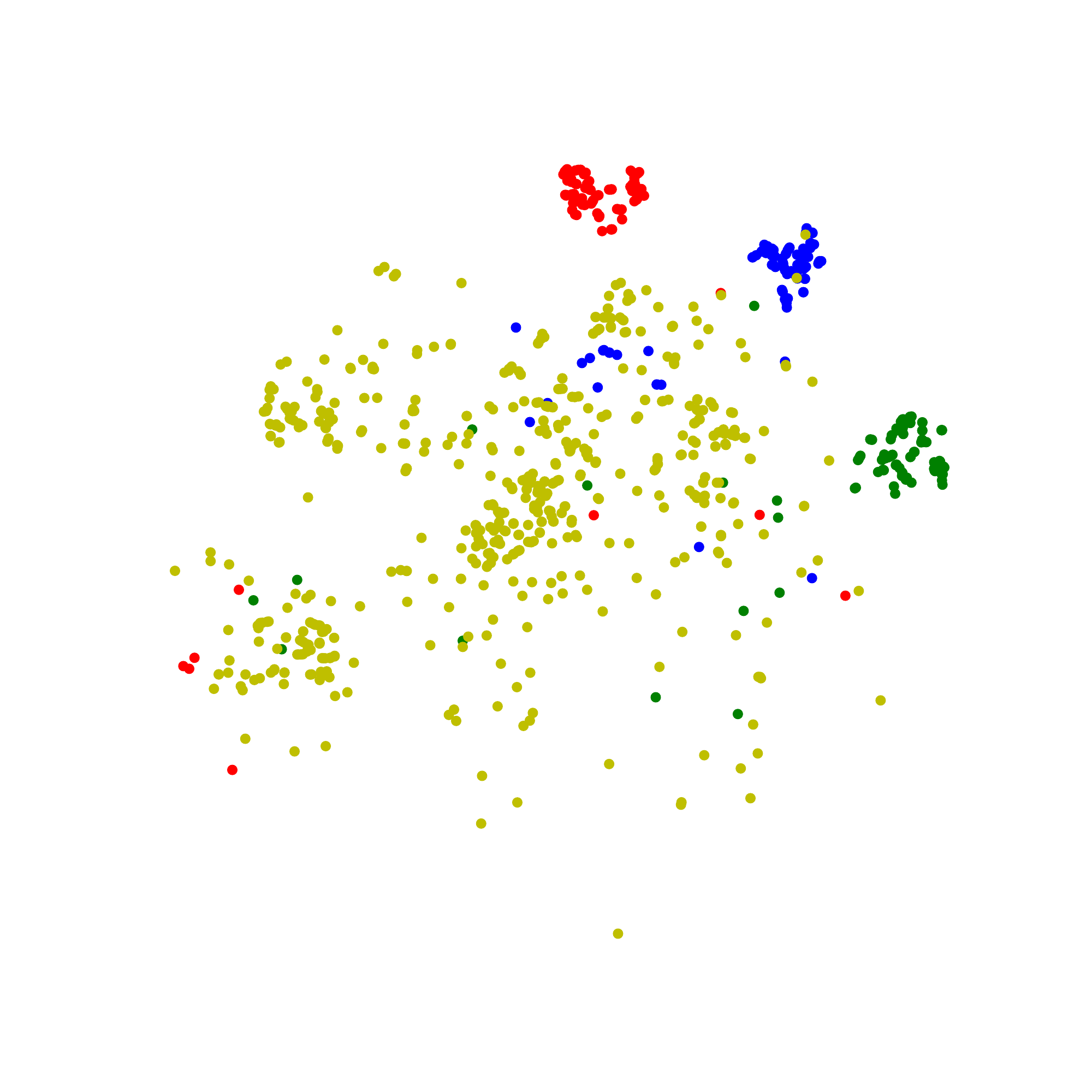}}
  \tikz{\draw[densely dotted](0, 3.5) -- (0,0);}
  \subfigure[Task 3]{\label{fig:conmatd}\includegraphics[height=3.5cm]{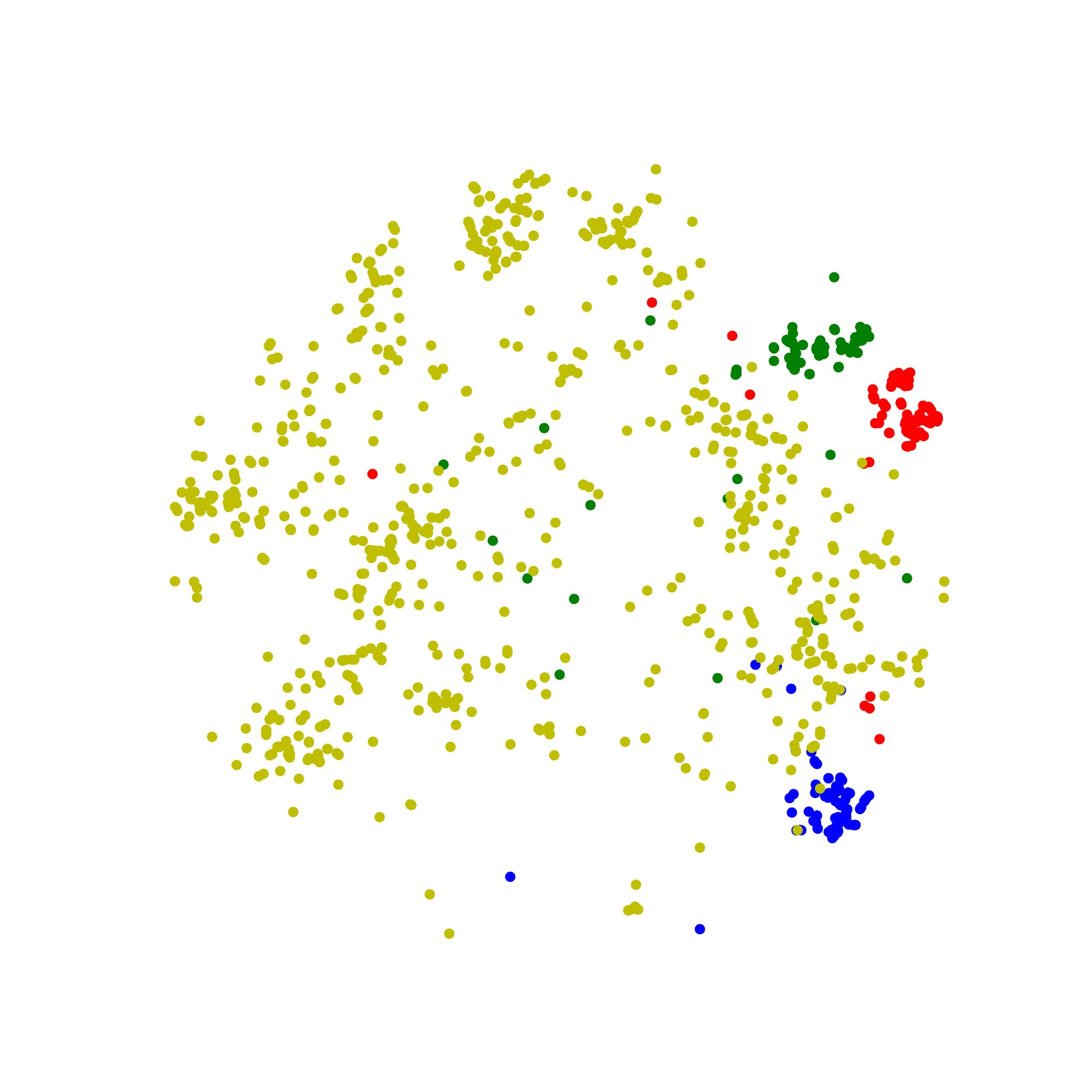}}
  \tikz{\draw[solid](0, 3.5) -- (0,0);}
  \subfigure[Confusion matrix]{\label{fig:conmate}\includegraphics[height=3.5cm]{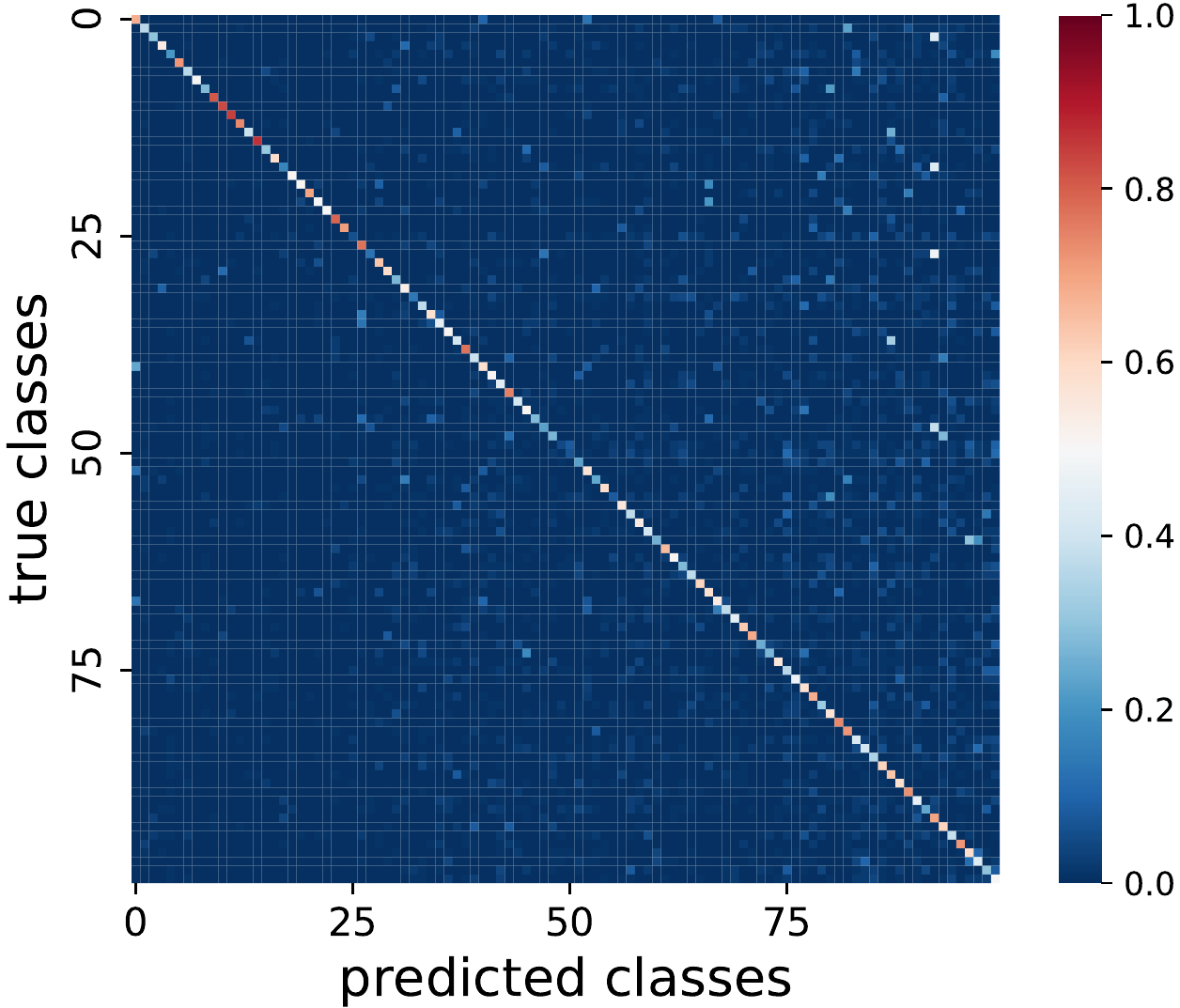}}
  }
  \caption{Visualization results of eTag on CIFAR-100 with $4$ evenly divided tasks. (a)-(d) show $3$ of $25$ classes in each task. (e) demonstrates the final confusion matrix.}
  \label{fig:conmat} 
\end{figure*}

\paragraph{Visualization Results.}
We qualitatively plot the t-SNE results and confusion matrices for Fine, Joint, GFR, and eTag on $4$ evenly divided tasks of CIFAR-100. For better visualization, $3$ out of $25$ classes for each task are randomly selected for t-SNE. Figs.\ref{fig:conmata}-\ref{fig:conmatd} show that eTag reliably retains the learned classes, keeping the old classes apart from the new ones. The final confusion matrix in Fig.\ref{fig:conmate} demonstrates that eTag alleviates the class imbalance problem.
Please see Appendix~\ref{app:a3} for more visualization results\footnote{Time consumption and comparisons with different sample generative methods are given in Appendix~\ref{app:a4} and \ref{app:a5}, respectively.}.
\begin{figure}[!t]
\begin{minipage}[b]{0.49\linewidth}
  \centering
  \includegraphics[width=\linewidth]{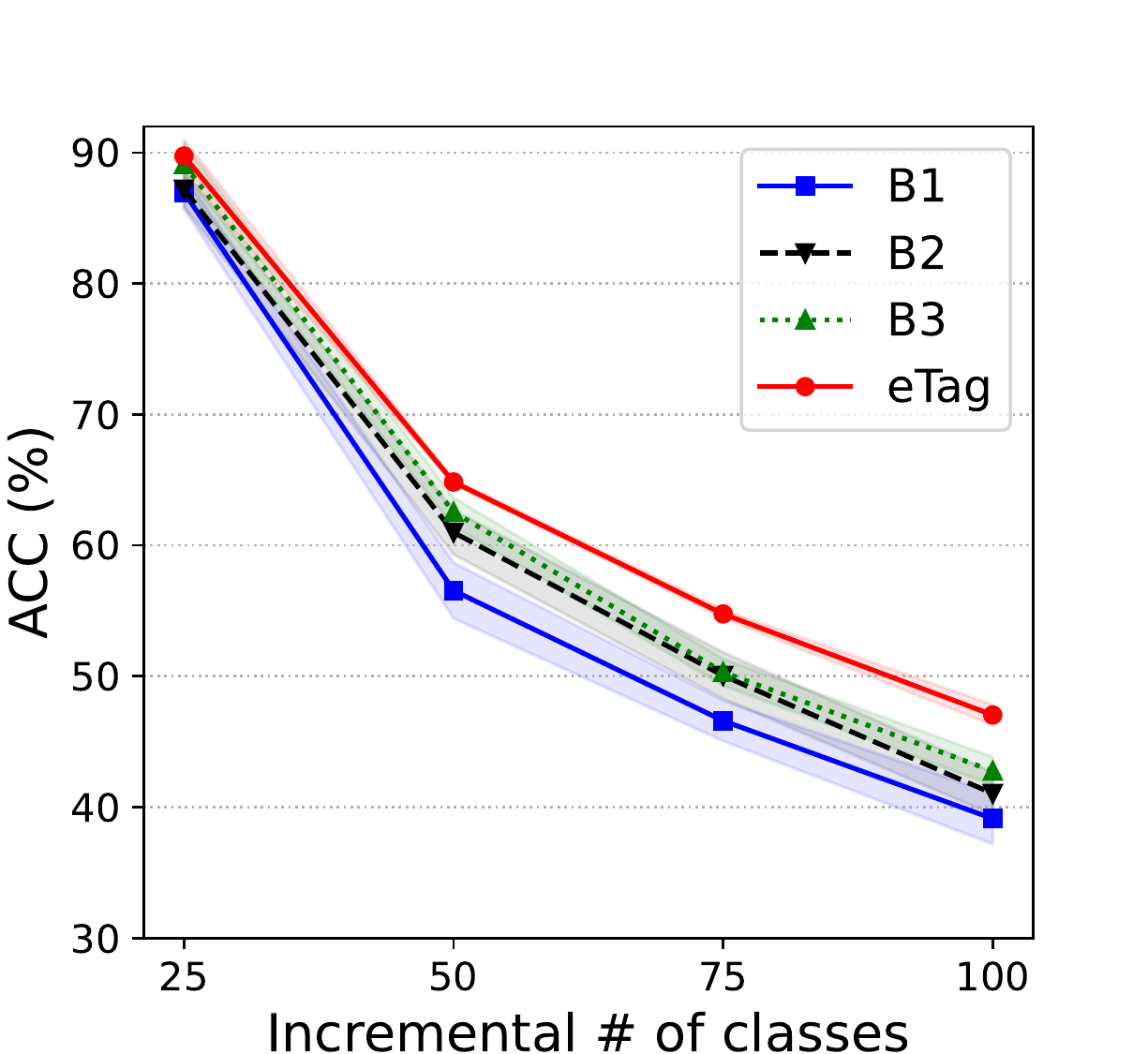}
  \caption{Component's effect}
  \label{fig:comeff}
\end{minipage}
\begin{minipage}[b]{0.49\linewidth}
  \centering
  \includegraphics[width=\linewidth]{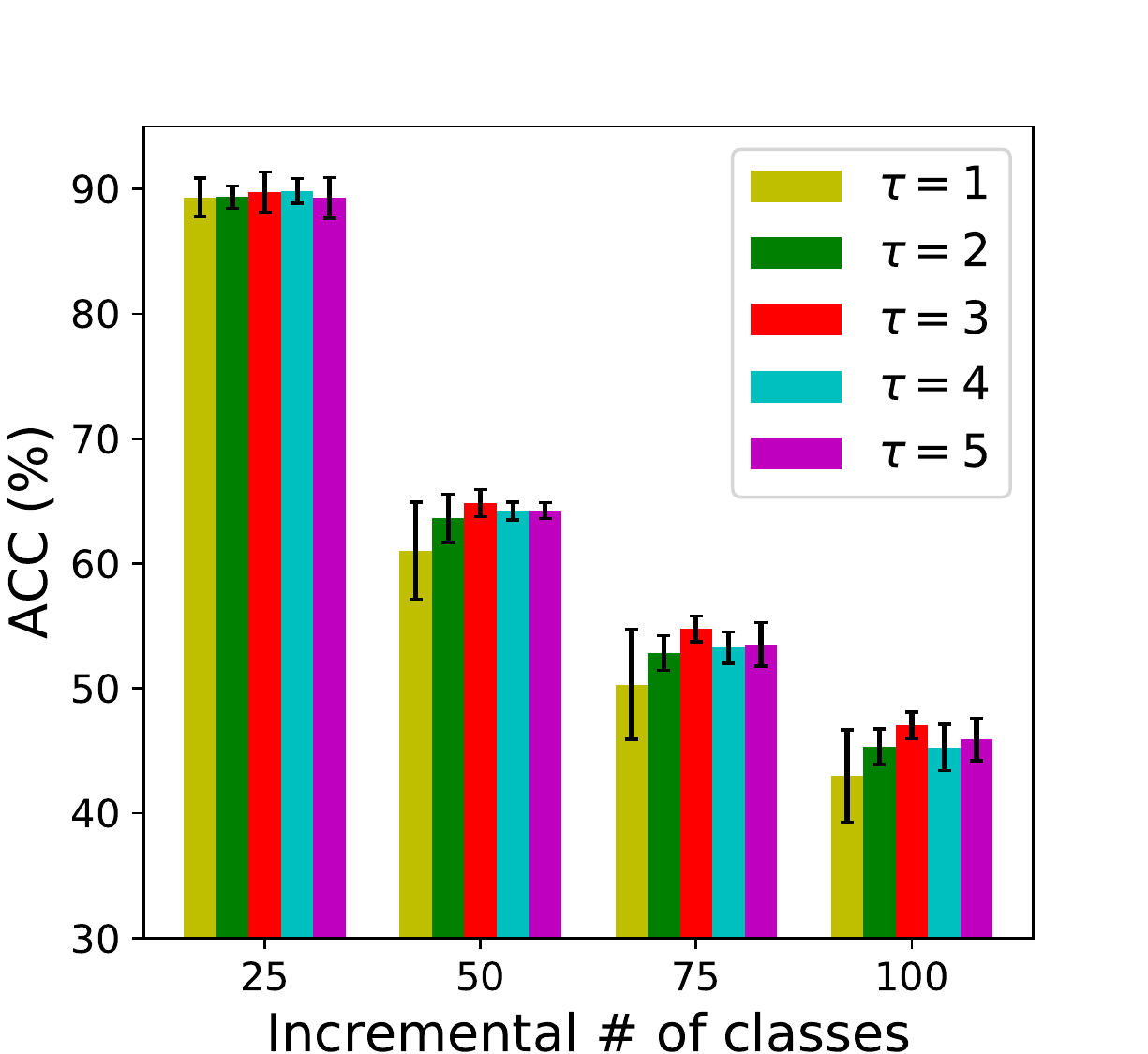}
  \caption{Parameter's effect}
  \label{fig:pareff}
\end{minipage}
\end{figure}

\begin{table}[!t]
  \caption{Ablation study results. Generation types ($GE$): sample generation ($S$), na\"ive ($N$) / task-oriented ($T$) feature generation; knowledge distillation ($KD$) types: na\"ive ($N$) / embedding ($E$) KD; Whether use $SS$ task / data-buffer ($DB$).}
  \label{tab:2}
  \centering
  \resizebox{.998\linewidth}{!}{
    \begin{tabular}{c|cc|ccccc}
    \toprule
    Method & PodNet & ABD & B0 & B1 & B2 & B3 & eTag \\
    \midrule
    $GE$  & \xmark & $S$ & $T$ & $N$ & $T$ & $N$ & $T$ \\
    $KD$  & $E$ & $N$ & $E$ & $N$ & $N$ & $E$ & $E$ \\
    $SS$  & \xmark & \xmark & \xmark & \xmark & \xmark & \cmark & \cmark \\
    $DB$  & \cmark & \xmark & \xmark & \xmark & \xmark & \xmark & \xmark \\
    \midrule
    $A (\uparrow)$	& 58.05 & 57.31 & 58.38 & 57.30 & 59.82 & 61.18 & 64.10 \\
    \bottomrule
    \end{tabular} }
\end{table}

\subsection{Ablation Studies} \label{subsec:abl}
\paragraph{Effects of Each Component.} 
We conduct experiments with three baselines on $4$ CIL tasks with equal classes of CIFAR-100. In Fig.\ref{fig:comeff}, B1 means na\"ive distillation and na\"ive feature generation. B2 replaces B1's na\"ive feature generation with the task-oriented generation, while B3 replaces B1's na\"ive distillation with embedding distillation. Fig.\ref{fig:comeff} shows that embedding distillation (B3) promotes the performance on the first task and maintains a similar trend to B1 in the CIL process, while task-oriented generation (B2) improves overall performance by reducing degradation. eTag combines their merits, accomplishing the best performance.

\paragraph{Effects of SS.}
We analyze the effects of SS tasks. In Tab.\ref{tab:2}, B0 does not use the self-supervised task in eTag. Podnet~\citep{douillard2020podnet} and ABD~\citep{smith2021always} are also comparisons\footnote{Podnet transfers vanilla feature maps following FitNet~\citep{romero2014fitnets}, while eTag develops SS-augmented embeddings driven by a meaningful SS task, complementary to the primary supervised task with richer feature knowledge. ABD is a recently proposed data-free CIL method in line with our eTag.}. Overall we see in Tab.\ref{tab:2} the efficacy of SS. Regarding data privacy and security issues, Podnet has to store exemplars from previous tasks, while our eTag has no need. eTag also performs better than ABD. Since there is no exemplar, learning from the history network is the primary way to prevent the model from forgetting. 
We also compared the effects of different SS tasks. In Tab.\ref{tab:difssl}, compared with jigsaw puzzles~\citep{noroozi2016unsupervised} and color channel permutation~\citep{zhang2016colorful}, the engaged random rotations help transfer more dark knowledge from history networks to the current one. Without SS, the current network cannot acquire sufficient history knowledge, resulting in a significant drop in performance from $64.10\%$ (eTag) to $58.38\%$ (B0).

\begin{table}[!t]
  \caption{Effects of different SS augmented tasks.}
  \label{tab:difssl}
  \centering
    \begin{tabular}{cccc}
    \toprule
    \makecell[c]{No \\ SS} & \makecell[c]{jigsaw \\ puzzles} & \makecell[c]{color channel \\ permutation} & \makecell[c]{random \\ rotations} \\
    \midrule
    58.38 & 59.11 & 56.66 & 64.10 \\
    \bottomrule
    \end{tabular}
\end{table}

\paragraph{Parameter Analysis.} 
We study the sensitivity of the only manual parameter $\tau$ in Eq.\eqref{eq:4}. The experiments are carried out on $4$ CIL tasks with identical classes on CIFAR-100. In Fig.\ref{fig:pareff}, the overall eTag is insensitive to the values of $\tau$. The maximum ACCs (average ACC + std) for different values of $\tau$ on each incremental task are almost identical, although the standard deviation is relatively large when $\tau=1$. That means maintaining the identical embeddings before and after CIL causes the model to be less plastic to learn new information.

\section{Conclusions}
To embrace the merits of both incrementally training a feature extractor and estimating the feature distribution, we develop a class-incremental learning algorithm with \textit{e}mbedding distillation and \textit{Ta}sk-oriented \textit{g}eneration (eTag). eTag incrementally distills the embeddings from intermediate block outputs to gain more knowledge for the feature extractor, and endows generative networks to produce features that match the classifier. Extensive experiments on CIFAR-100 and ImageNet-sub verified the effectiveness of our eTag in tracking the forgetting problem.

Albeit satisfactory performances, CIL is far from being solved. Particularly, eTag performs lower than the upper bound of joint training. We hope that the ideas and the success of eTag will inspire more effective CIL strategies to reduce the remaining performance gap. In the future, it is desirable to improve the training efficiency of eTag and engage the pre-trained big backbone model for CIL.


\clearpage
\bibliographystyle{named}
\bibliography{ijcai23}

\clearpage
\appendix
\renewcommand\thefigure{\thesection.\arabic{figure}} 
\setcounter{figure}{0}
\renewcommand\thetable{\thesection.\arabic{table}} 
\setcounter{table}{0}

\section{Appendix of ``eTag: Class-Incremental Learning with Embedding Distillation and Task-Oriented Generation"}
\begin{figure*}[!b]
    \centering
    \subfigure[5 incremental tasks]{\label{fig:3a}\includegraphics[height=4.4cm]{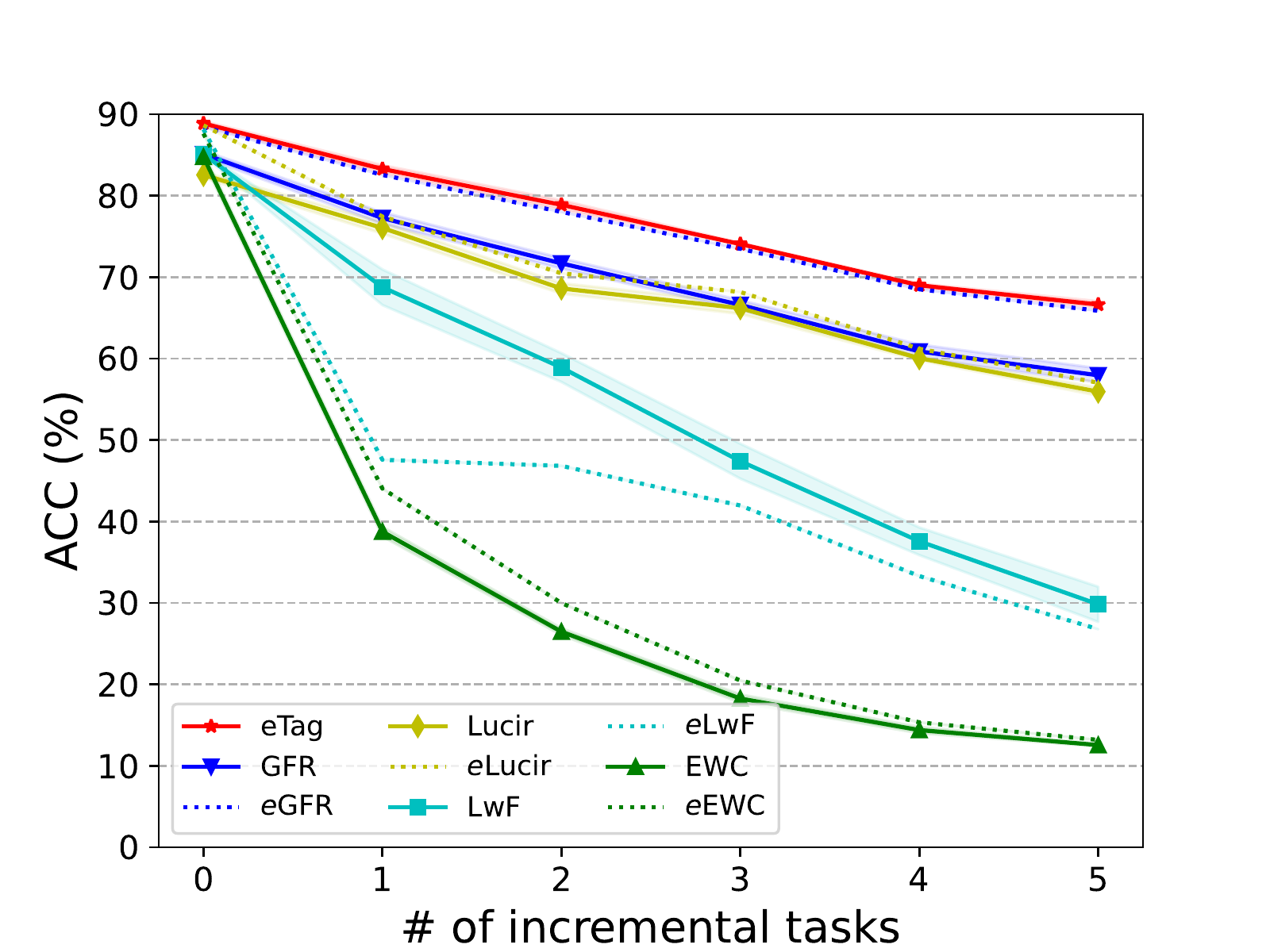}}
    \hspace*{\fill}
    \subfigure[10 incremental tasks]{\label{fig:3b}\includegraphics[height=4.4cm]{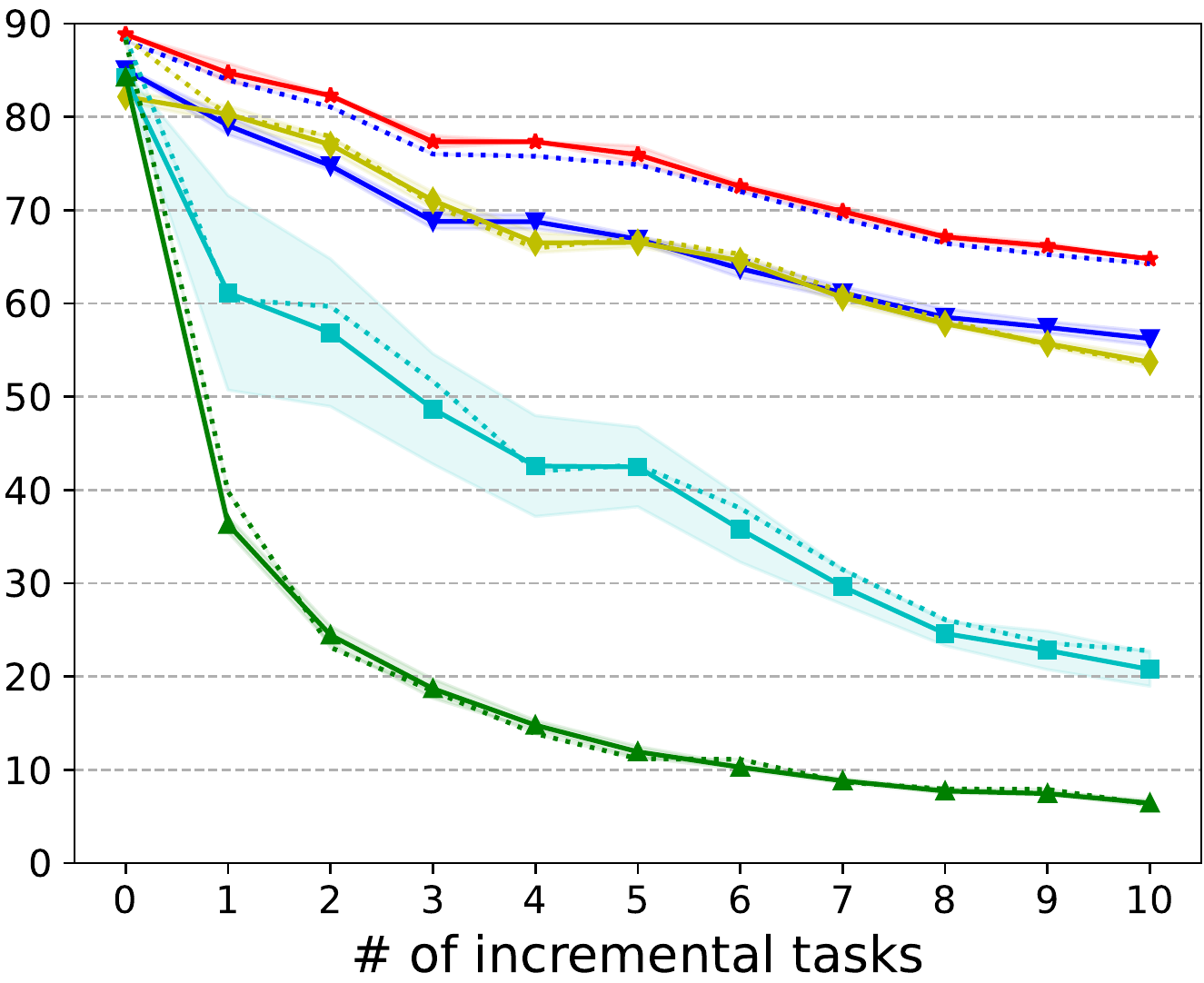}}
    \hspace*{\fill}
    \subfigure[25 incremental tasks]{\label{fig:3c}\includegraphics[height=4.4cm]{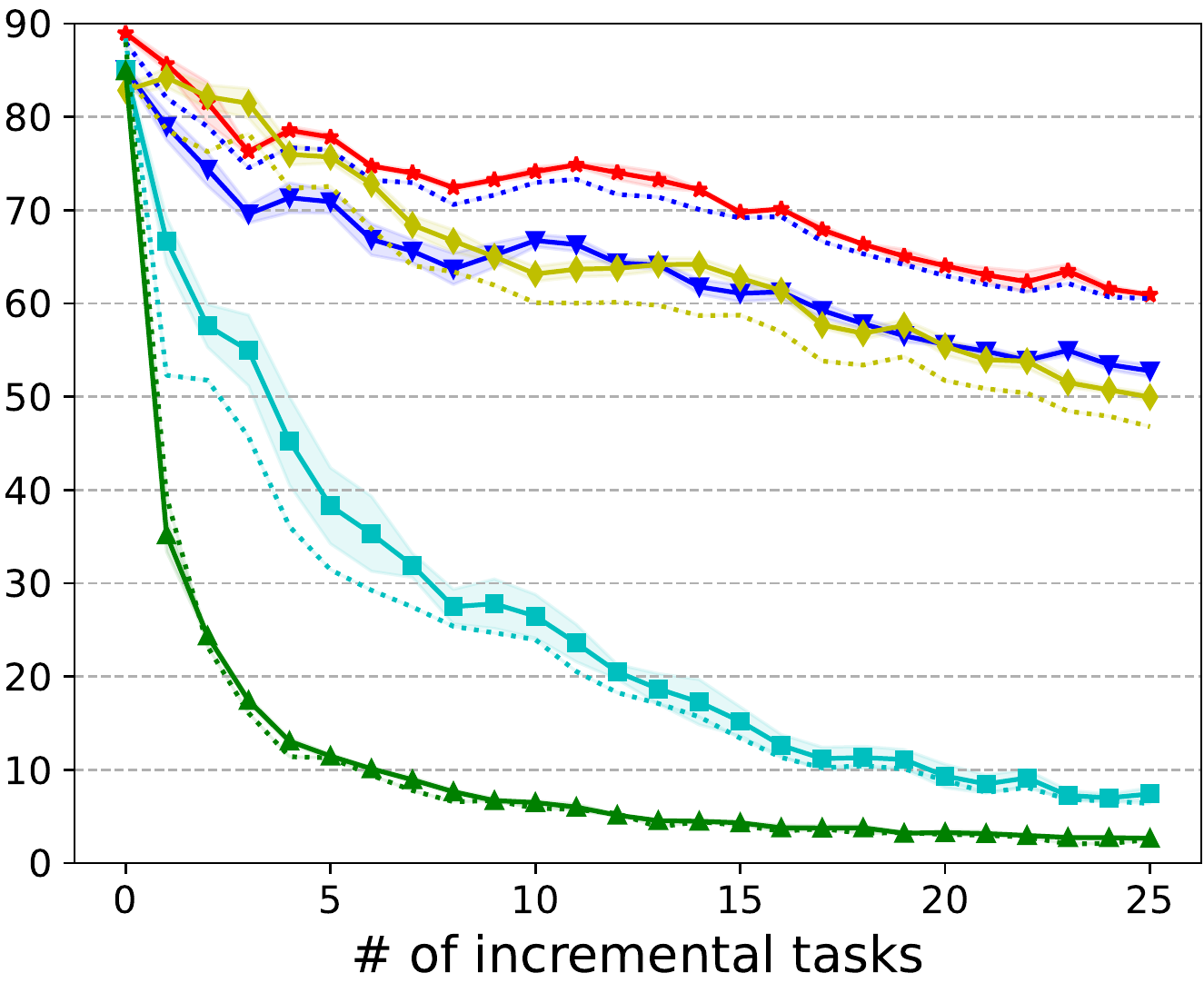}}
    \caption{Detailed ACCs on ImageNet-sub with $5$, $10$, and $25$ class-incremental tasks. Results are averaged after $3$ runs.}
    \label{fig:app1}
\end{figure*}

In this section, we provide further explanations about,
\begin{enumerate}[noitemsep,leftmargin=23.5pt]
    \item Implementation Details;
    \item Detailed ACCs on ImageNet-Sub;
    \item More Visualization Results;
    \item Time Consumption Analysis;
    \item Compared with Sample Generative Methods.
\end{enumerate}

\subsection{Implementation Details} \label{app:a1}
All models are built with the Pytorch framework~\citep{paszke2017automatic} and trained on V-100 GPUs. We modify the first convolutional layer of ResNet-18~\citep{he2016deep} with $3\times3$ kernels for CIFAR-100 and engage the original ResNet-18 for ImageNet-sub. The CIFAR-100 classification task is trained with $100$ epochs, ImageNet-sub with $70$, and the generator with $100$. We employ the Adam optimizer~\citep{kingma2015adam} with $128$ batch sizes, and train all models from scratch. The learning rates begin with $1e-3$ for classification and $1e-4$ for generator, and are divided by $10$ every $30$ epoch.

\subsection{Detailed ACCs on ImageNet-Sub} \label{app:2}
Fig.\ref{fig:app1} demonstrates the detailed ACCs on ImageNet-sub. Similar to the conclusion drawn in Sec.\ref{subsec:cileva} of the main paper, directly plugging in the embedding distillation in these SOTA methods is not constantly helpful for incremental tasks. For instance, on the $25$ incremental tasks of ImageNet-sub, $r$Lucir performs worse than Lucir.

\subsection{More Visualization Results} \label{app:a3}
We investigate Fine, Joint, GFR, and eTag further, and plot their t-SNE and the final confusion matrix in this section. All experiments are conducted on $4$ evenly divided tasks of CIFAR-100, and $3$ of $25$ classes of each task are randomly selected for better t-SNE visualization. 

\begin{figure*}[!hbp]
  \centering
  \stackunder[2pt]{\label{fig:6a}\includegraphics[height=4.3cm]{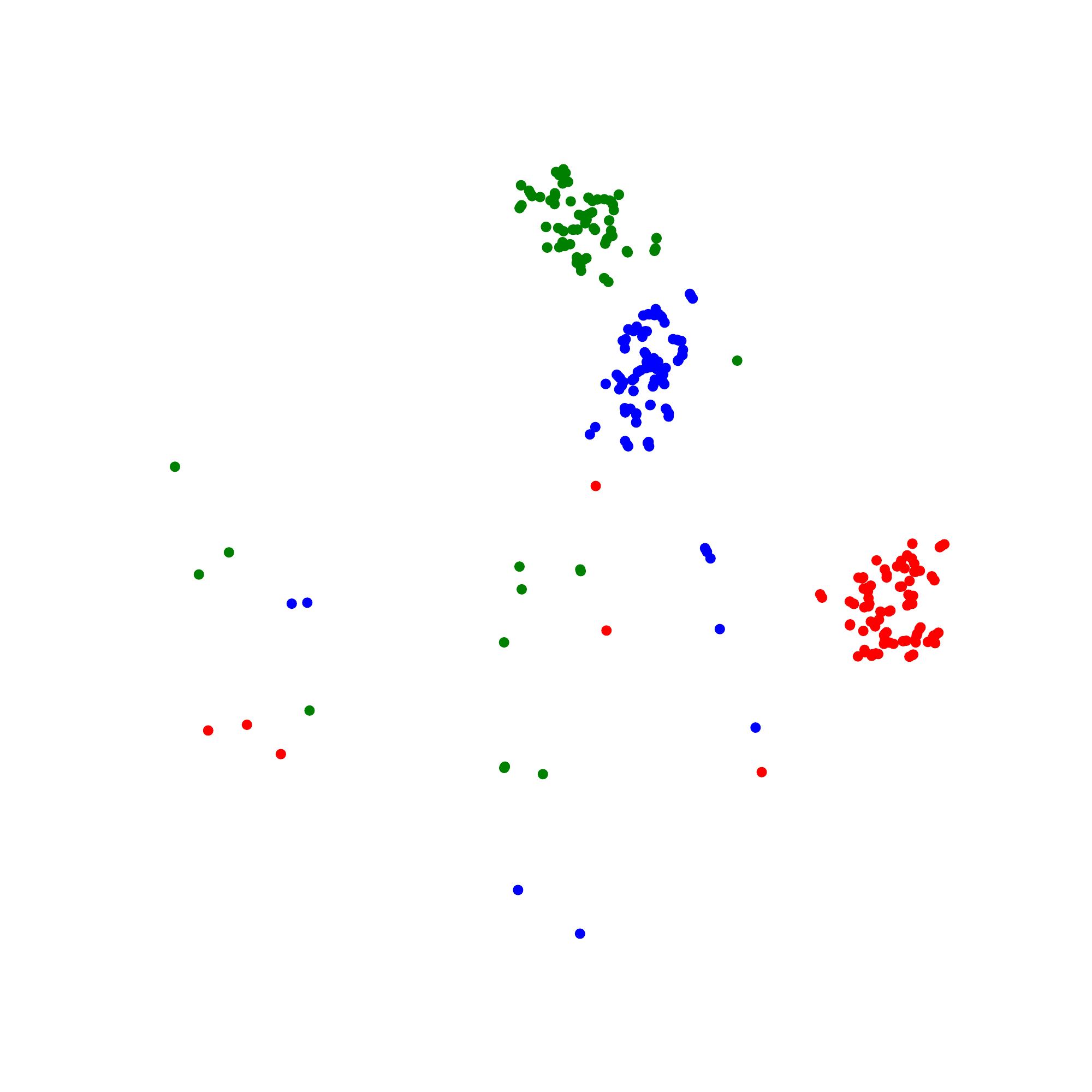}}{Task 0}
  \tikz{\draw[densely dotted](0, 4.3) -- (0,0);}
  \stackunder[2pt]{\label{fig:6b}\includegraphics[height=4.3cm]{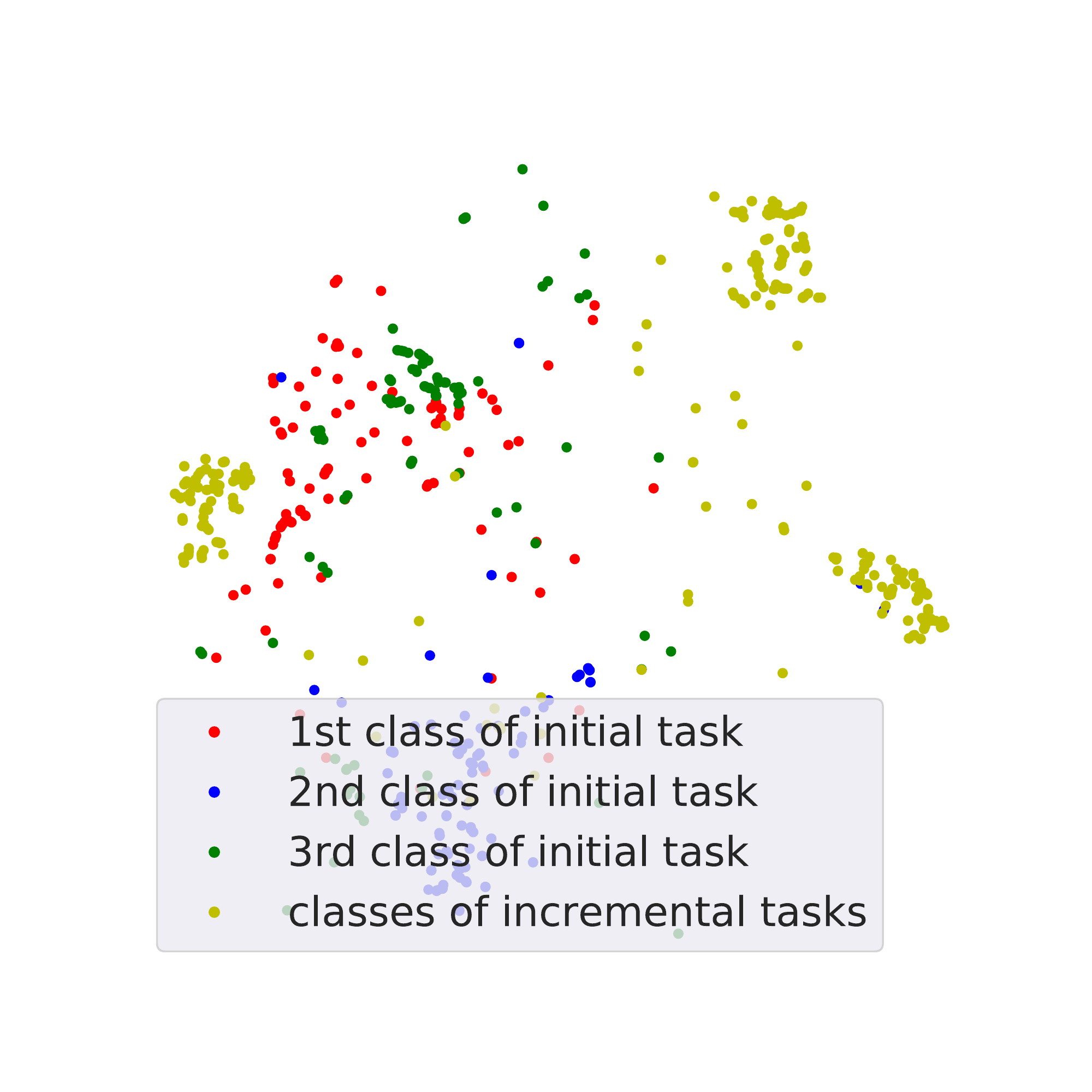}}{Task 1}
  \tikz{\draw[densely dotted](0, 4.3) -- (0,0);}
  \stackunder[2pt]{\label{fig:6c}\includegraphics[height=4.3cm]{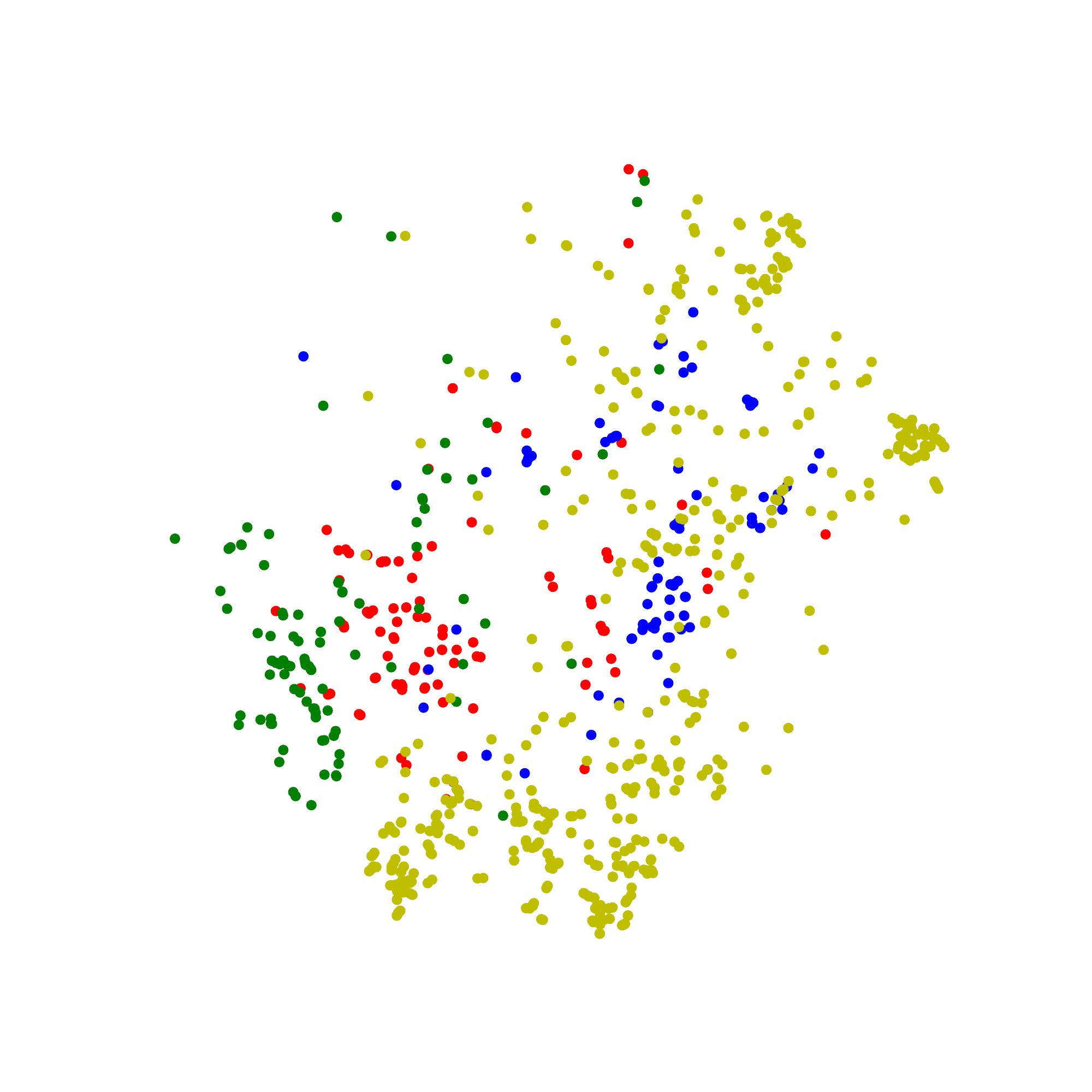}}{Task 2}
  \tikz{\draw[densely dotted](0, 4.3) -- (0,0);}
  \stackunder[2pt]{\label{fig:6d}\includegraphics[height=4.3cm]{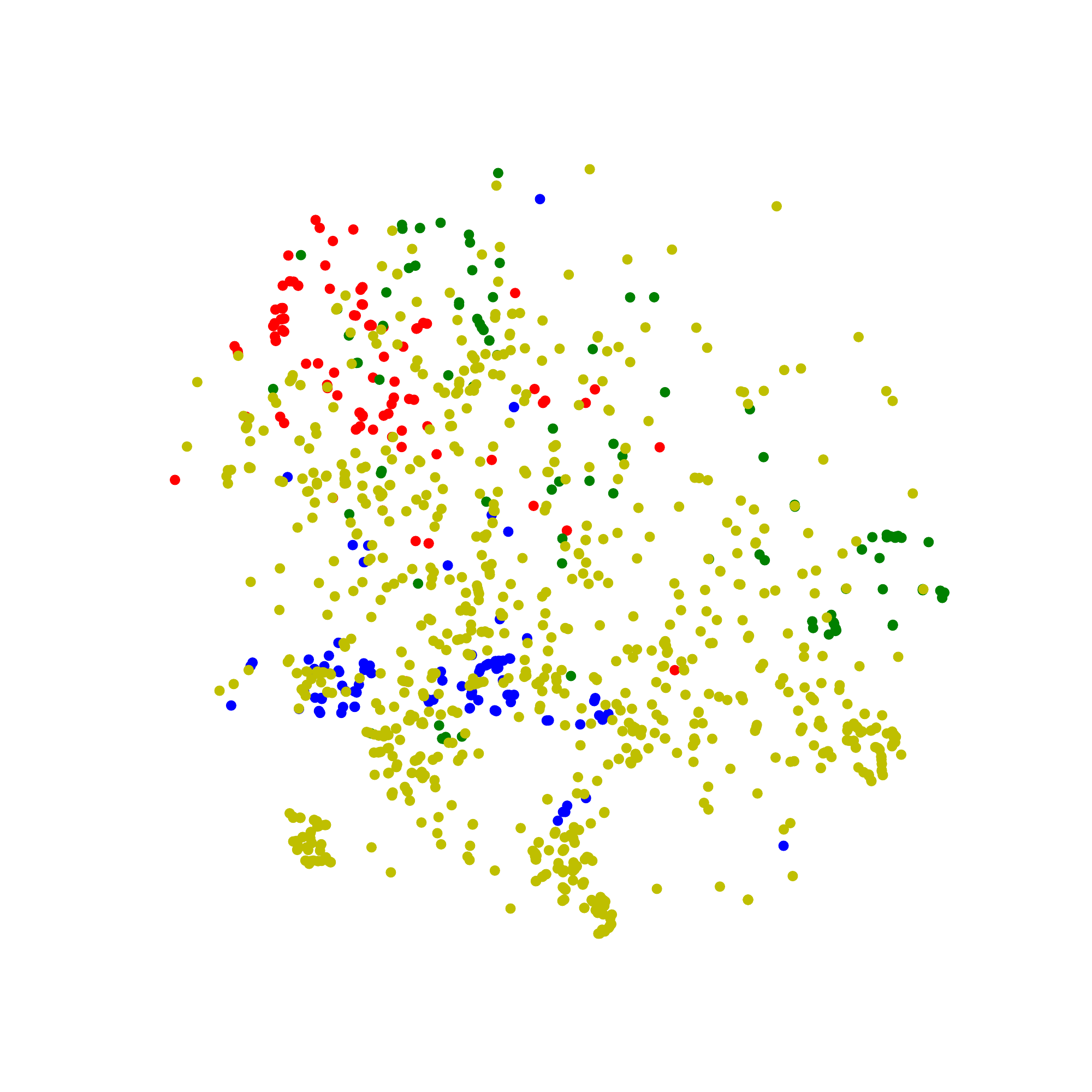}}{Task 3}
  \caption{t-SNE results of \textbf{Fine} on CIFAR-100 with 4 evenly divided tasks.}
  \label{fig:a2} 
\end{figure*}

\begin{figure*}[!hbp]
  \centering
  \stackunder[2pt]{\label{fig:6a}\includegraphics[height=4.3cm]{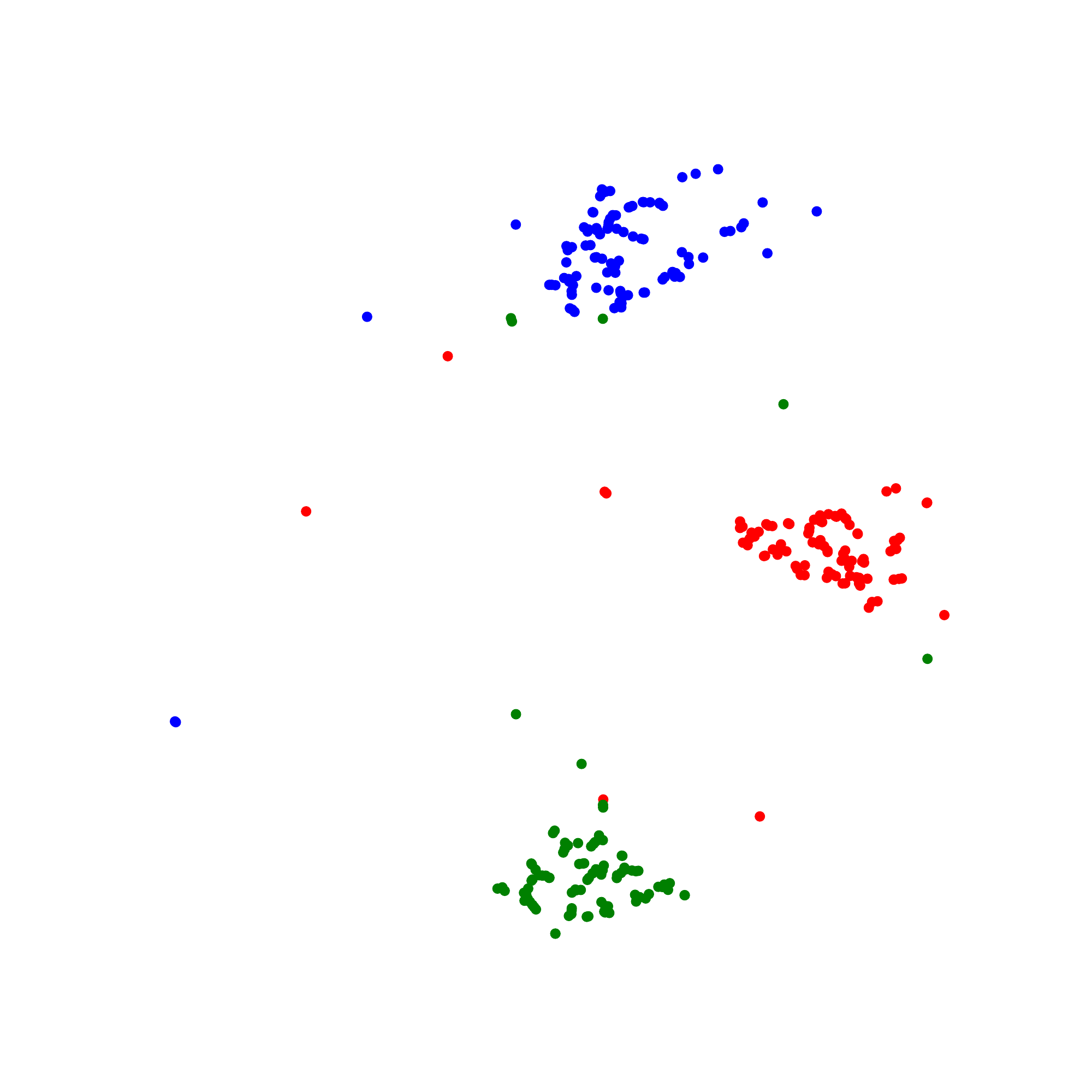}}{Task 0}
  \tikz{\draw[densely dotted](0, 4.3) -- (0,0);}
  \stackunder[2pt]{\label{fig:6b}\includegraphics[height=4.3cm]{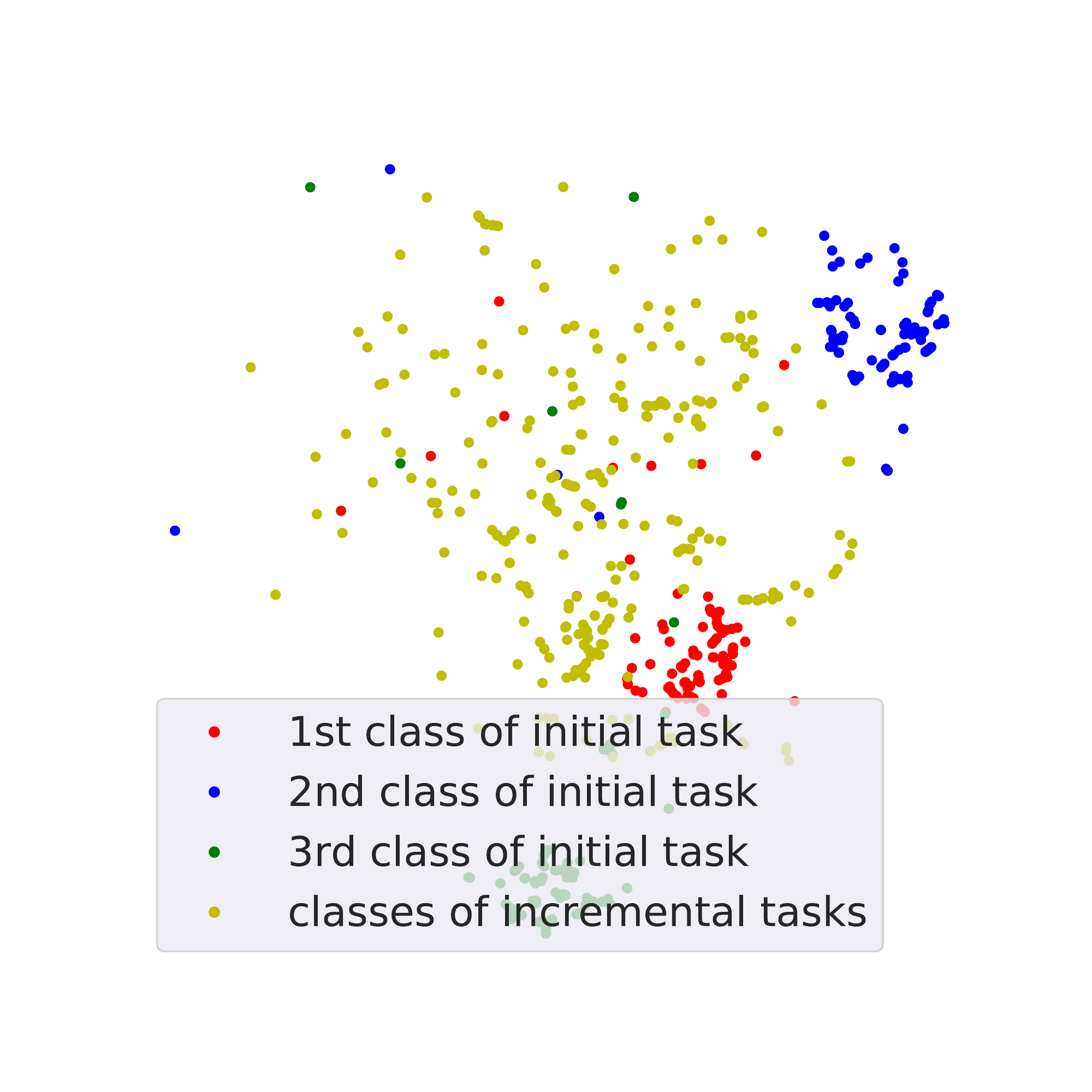}}{Task 1}
  \tikz{\draw[densely dotted](0, 4.3) -- (0,0);}
  \stackunder[2pt]{\label{fig:6c}\includegraphics[height=4.3cm]{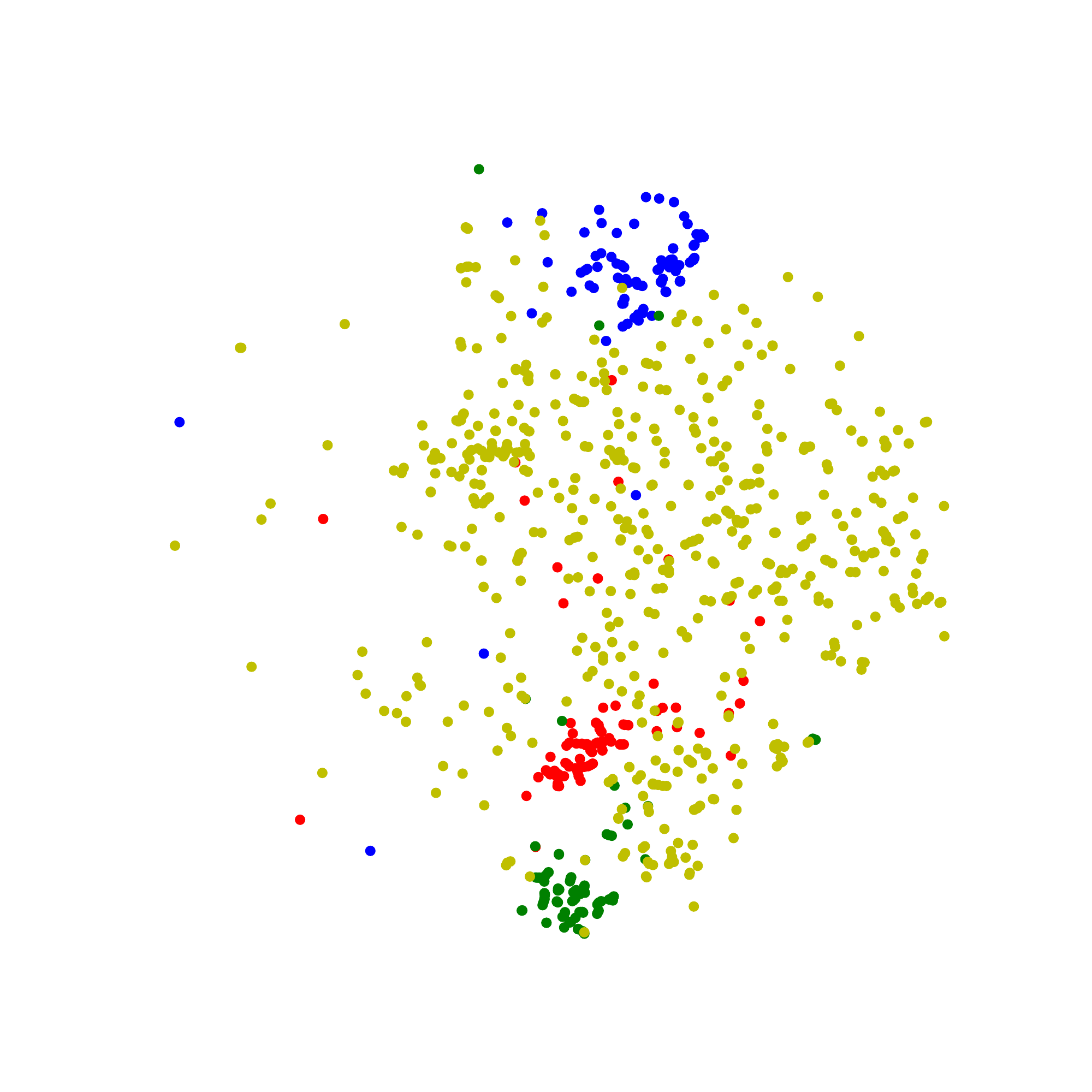}}{Task 2}
  \tikz{\draw[densely dotted](0, 4.3) -- (0,0);}
  \stackunder[2pt]{\label{fig:6d}\includegraphics[height=4.3cm]{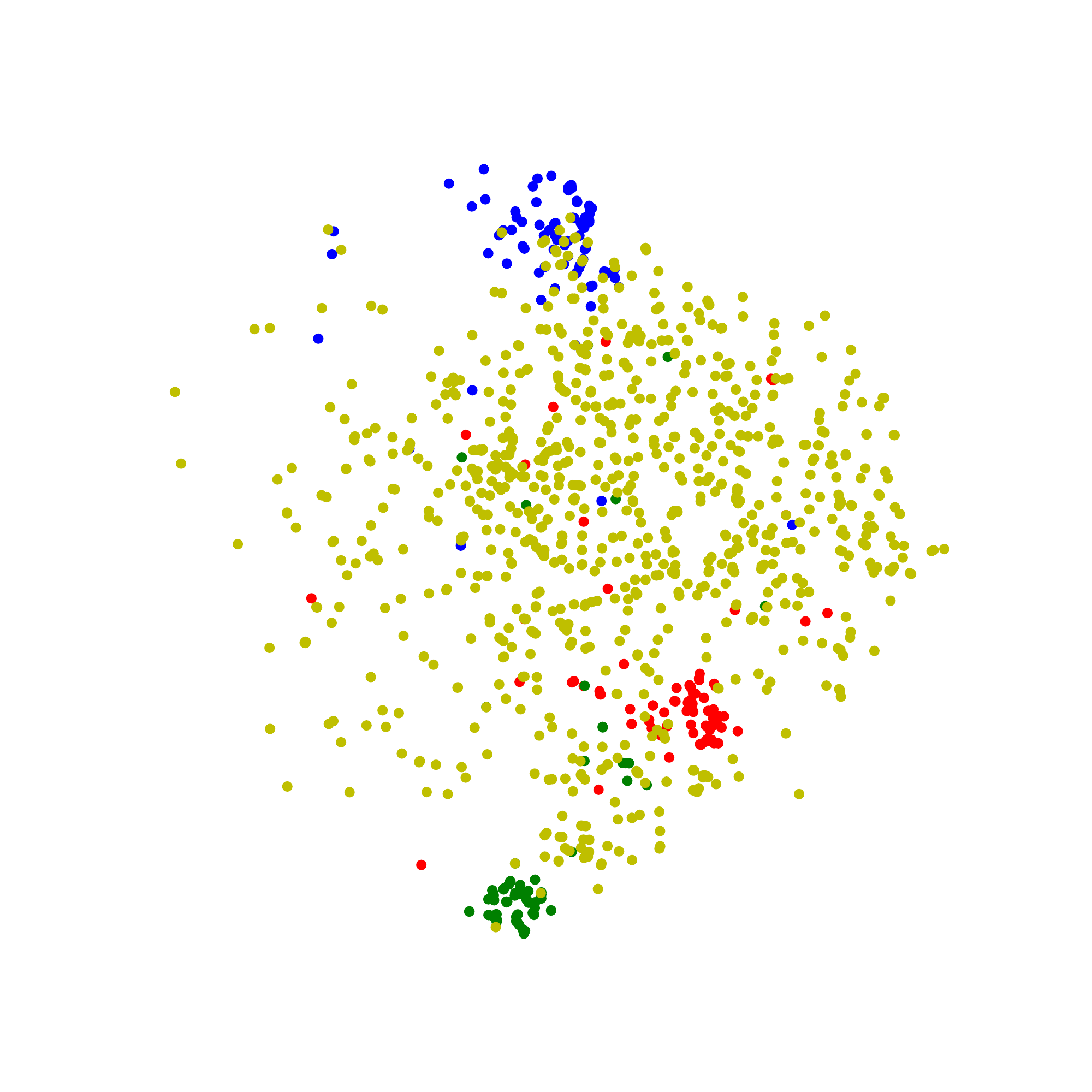}}{Task 3}
  \caption{t-SNE results of \textbf{GFR} on CIFAR-100 with 4 evenly divided tasks.}
  \label{fig:a3} 
\end{figure*}

\begin{figure*}[!hbp]
  \centering
  \stackunder[2pt]{\label{fig:6a}\includegraphics[height=4.3cm]{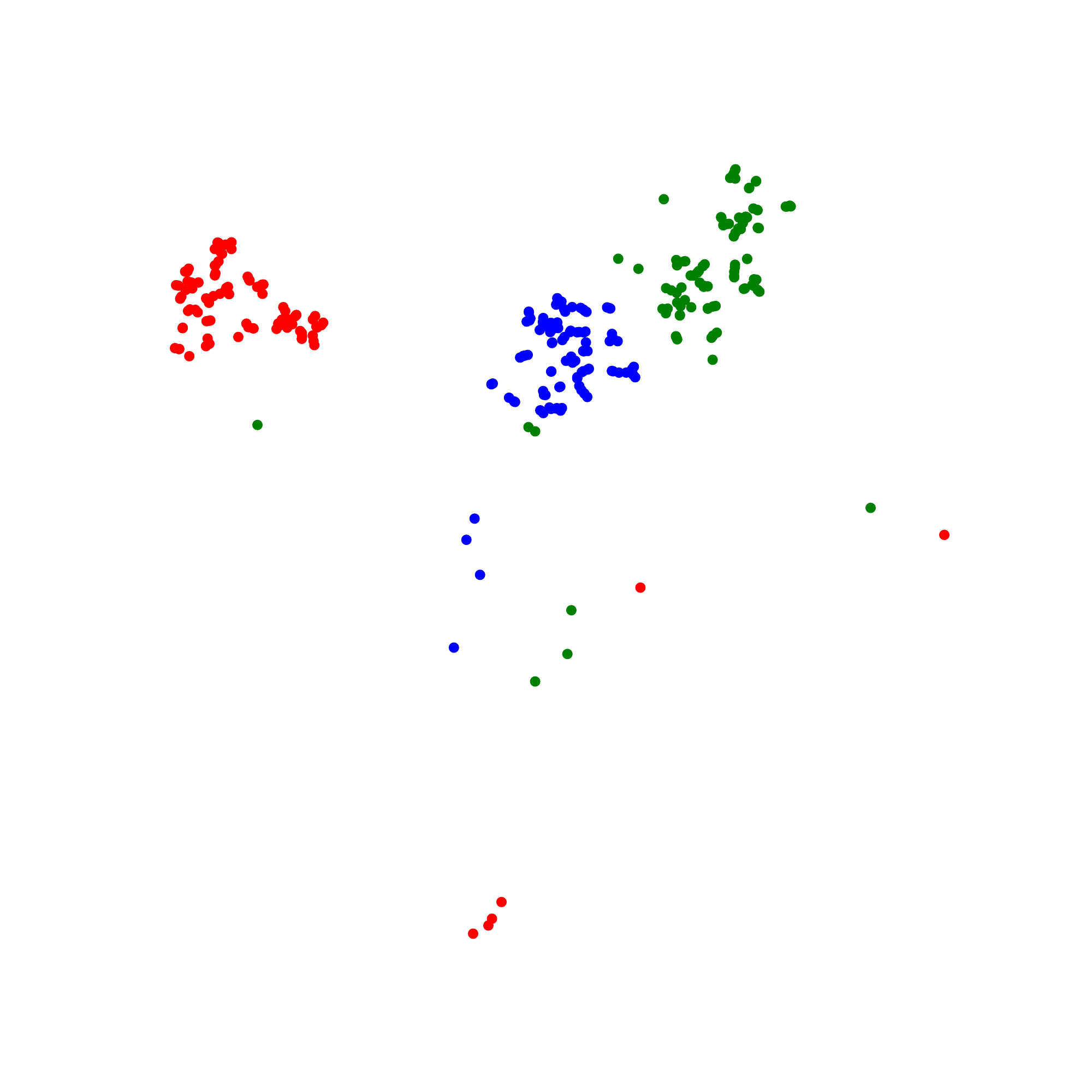}}{Task 0}
  \tikz{\draw[densely dotted](0, 4.3) -- (0,0);}
  \stackunder[2pt]{\label{fig:6b}\includegraphics[height=4.3cm]{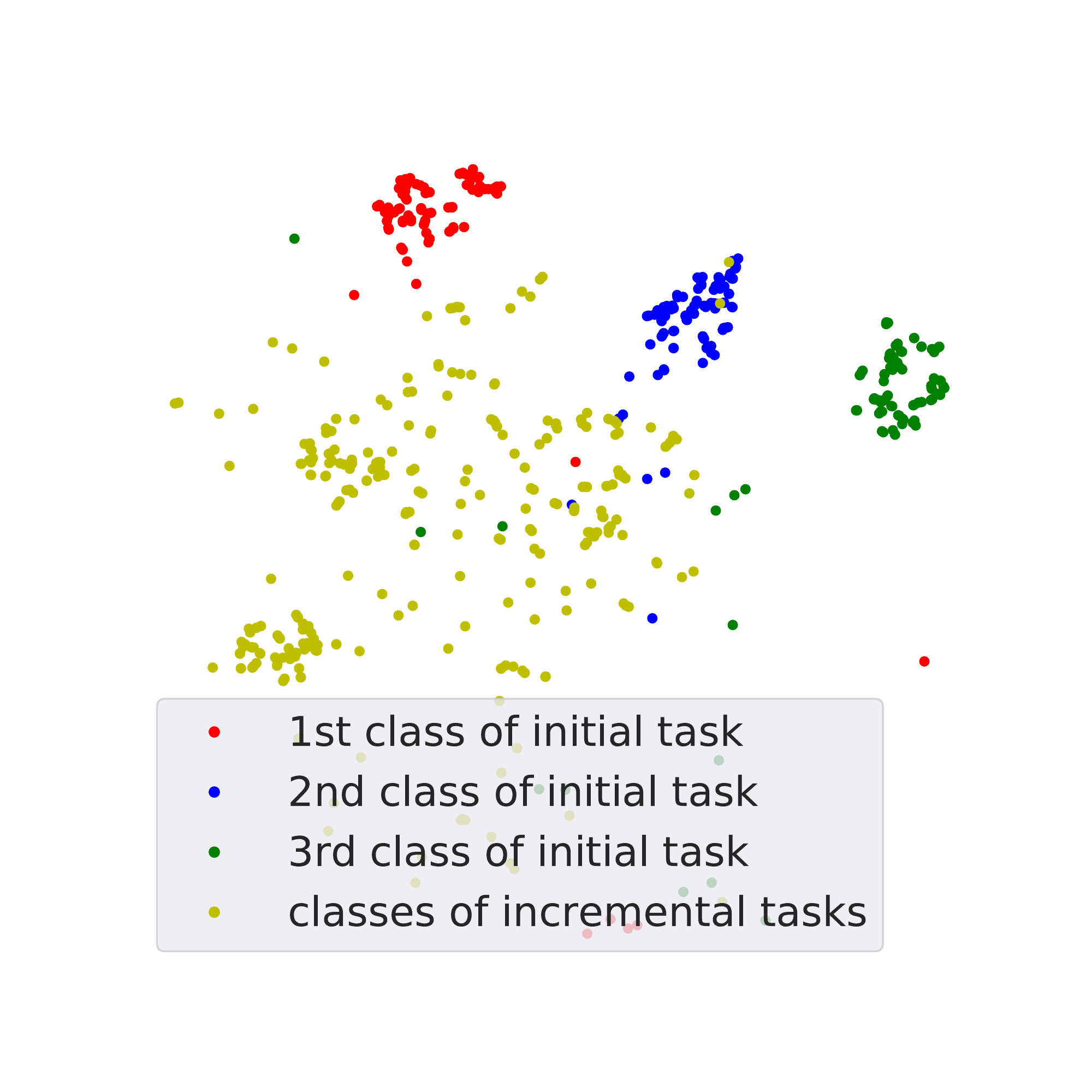}}{Task 1}
  \tikz{\draw[densely dotted](0, 4.3) -- (0,0);}
  \stackunder[2pt]{\label{fig:6c}\includegraphics[height=4.3cm]{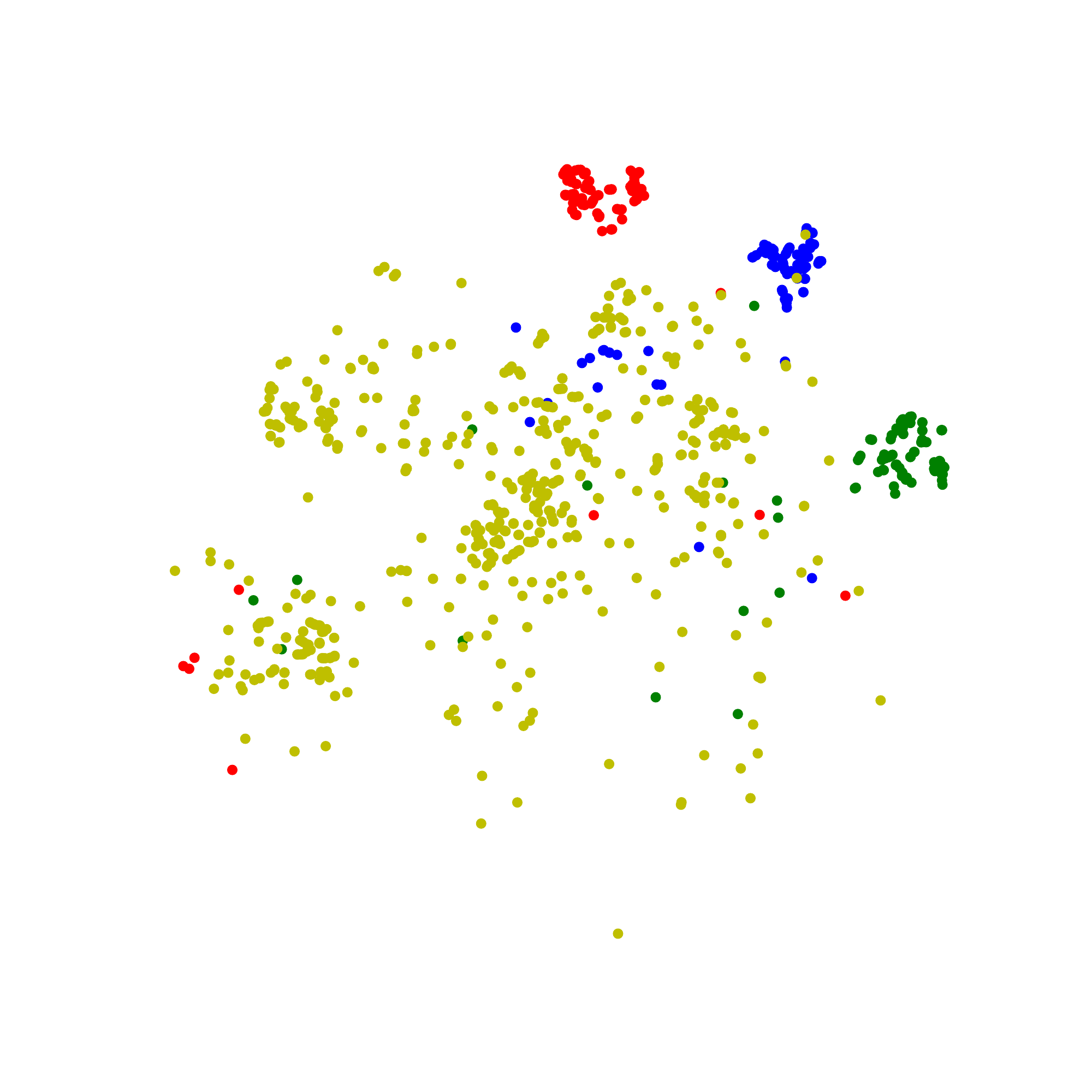}}{Task 2}
  \tikz{\draw[densely dotted](0, 4.3) -- (0,0);}
  \stackunder[2pt]{\label{fig:6d}\includegraphics[height=4.3cm]{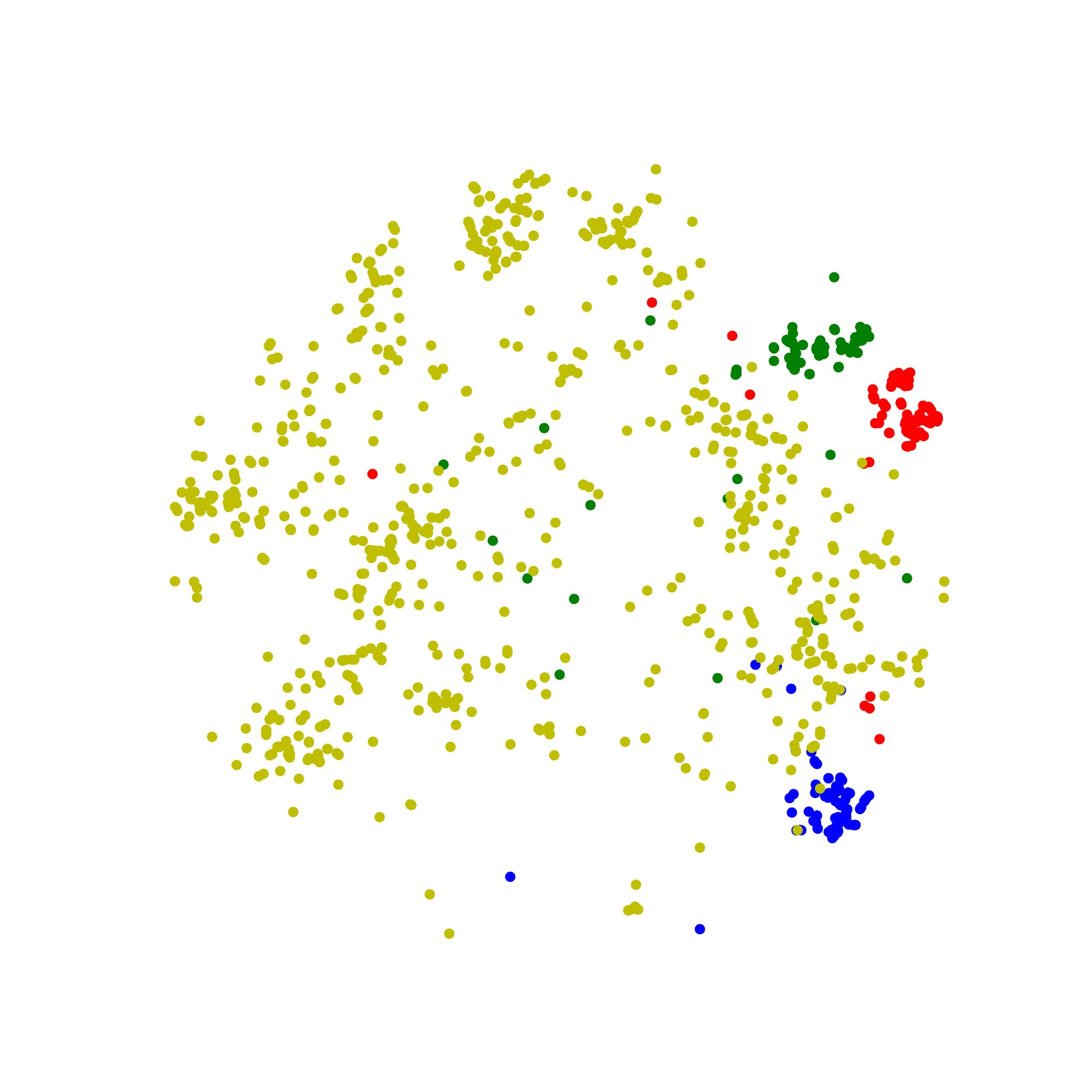}}{Task 3}
  \caption{t-SNE results of \textbf{eTag} on CIFAR-100 with 4 evenly divided tasks.}
  \label{fig:a4} 
\end{figure*}

\begin{figure*}[!hbp]
  \centering
  \stackunder[2pt]{\label{fig:6a}\includegraphics[height=4.3cm]{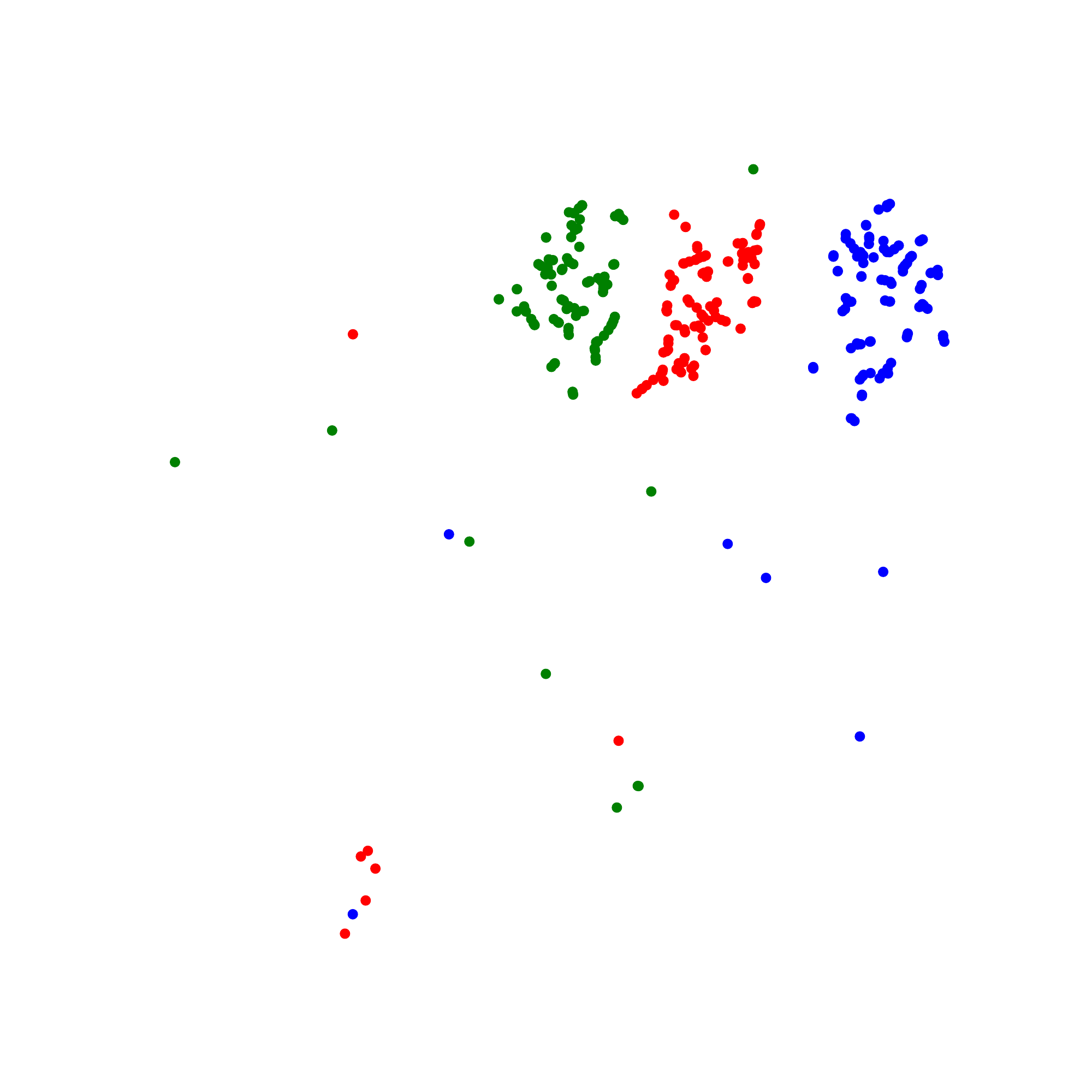}}{Task 0}
  \tikz{\draw[densely dotted](0, 4.3) -- (0,0);}
  \stackunder[2pt]{\label{fig:6b}\includegraphics[height=4.3cm]{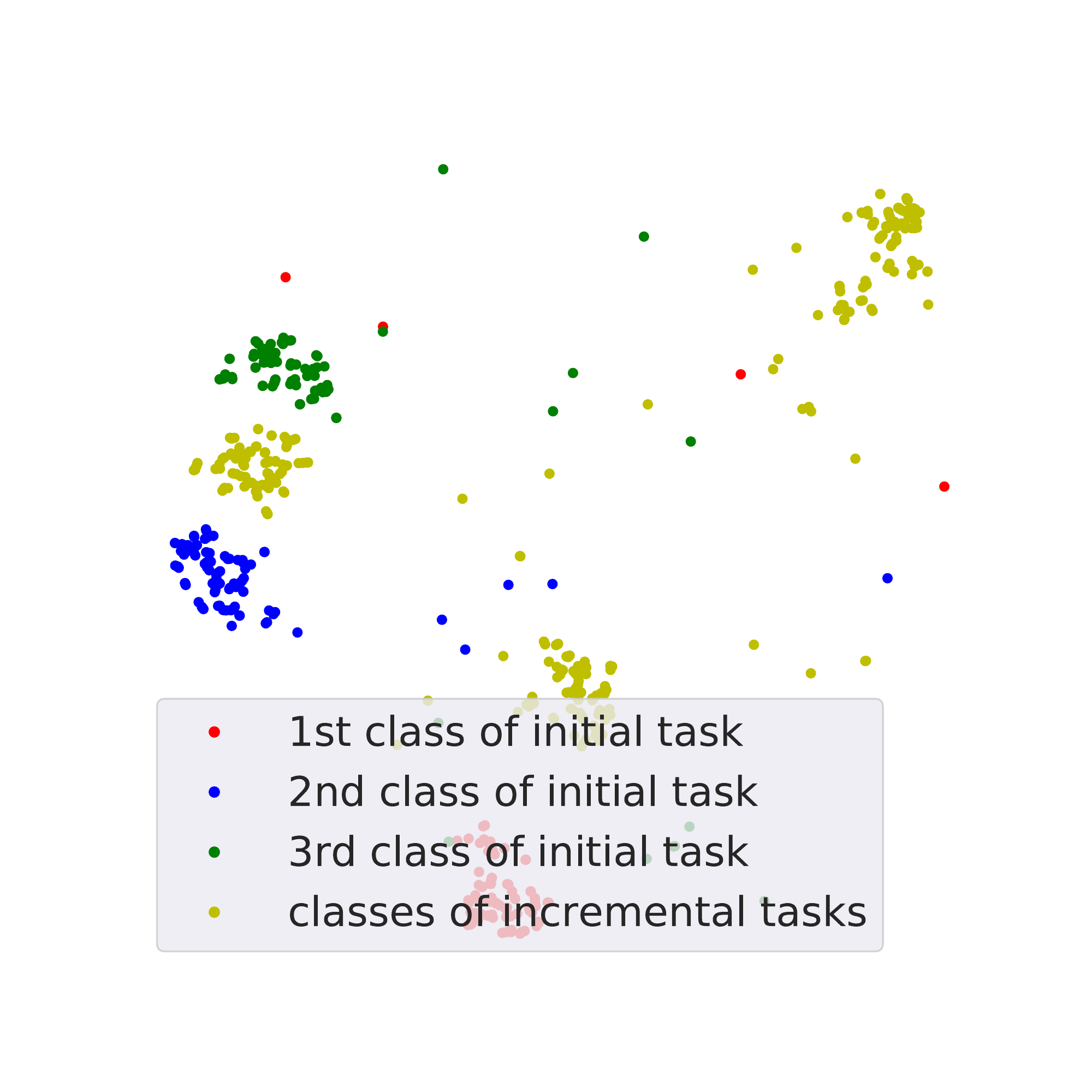}}{Task 1}
  \tikz{\draw[densely dotted](0, 4.3) -- (0,0);}
  \stackunder[2pt]{\label{fig:6c}\includegraphics[height=4.3cm]{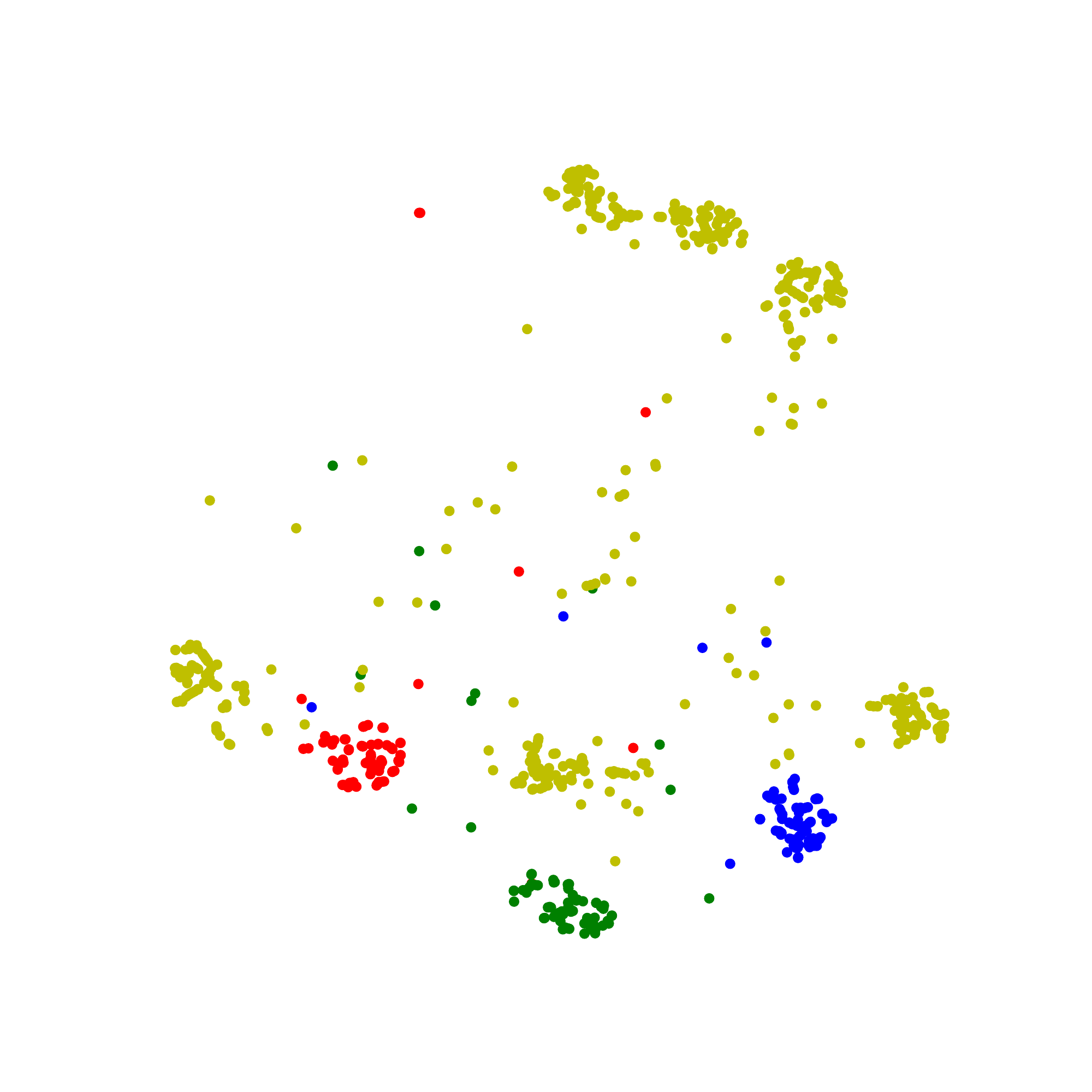}}{Task 2}
  \tikz{\draw[densely dotted](0, 4.3) -- (0,0);}
  \stackunder[2pt]{\label{fig:6d}\includegraphics[height=4.3cm]{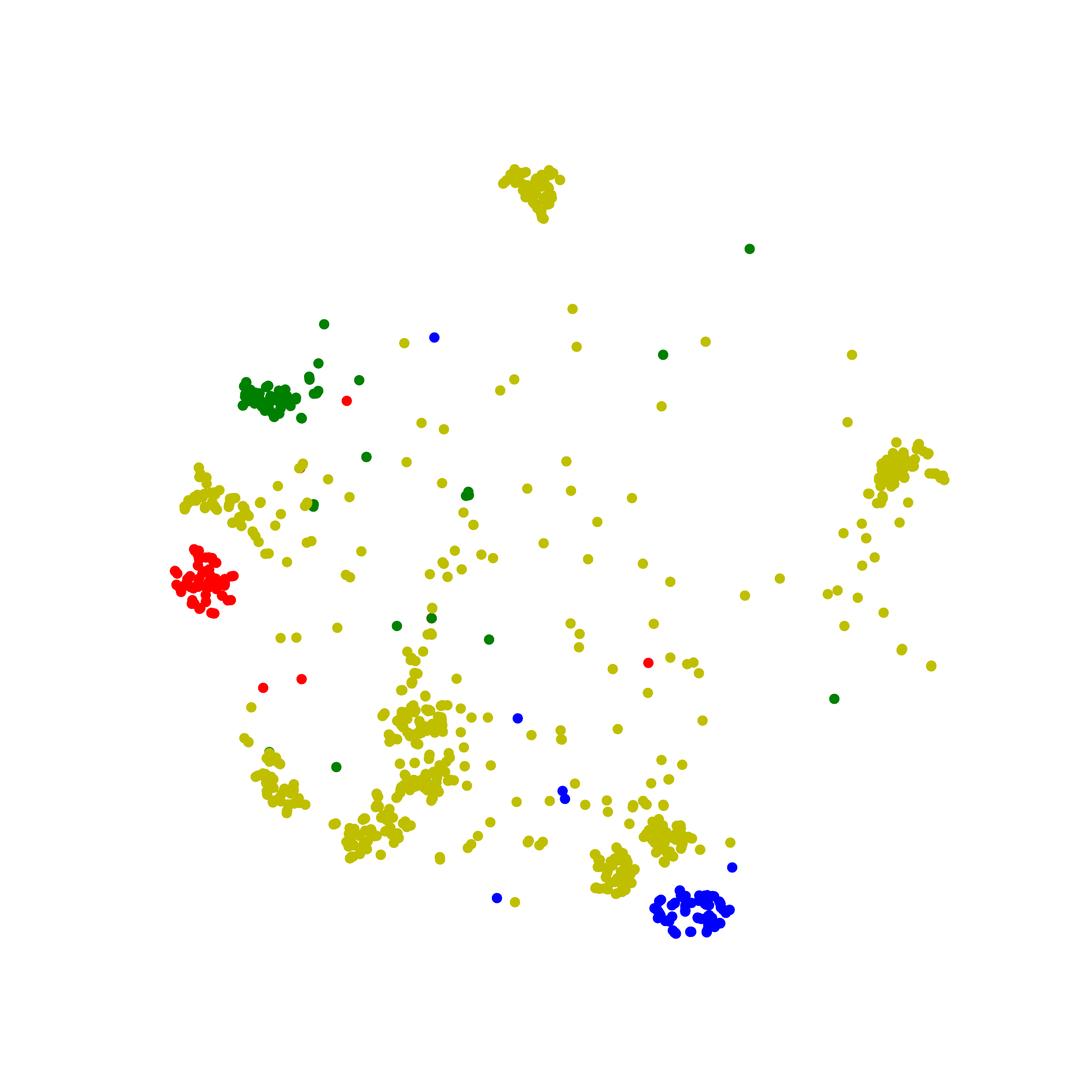}}{Task 3}
  \caption{t-SNE results of \textbf{Joint} on CIFAR-100 with 4 evenly divided tasks.}
  \label{fig:a5} 
\end{figure*}

Figs.\ref{fig:a2}-\ref{fig:a5} show that our proposed eTag achieves comparable t-SNE visualization results. Although all methods could separate classes well on the initial task, they show different results as these incremental tasks arrive. Among them, fine-tuning (Fine) the model completely confuses the previously-learned classes with the currently learned ones. GFR improves on Fine's results but still confounds the 3rd class learned in the initial task (red) with newly learned classes after the model learns $4$ tasks. In contrast, eTag reliably retains the learned classes and obtains comparable t-SNE results as that of joint training.

\begin{figure*}[!hbp]
    \centering
    \subfigure[Fine]{\label{fig:6a}\includegraphics[height=4.3cm]{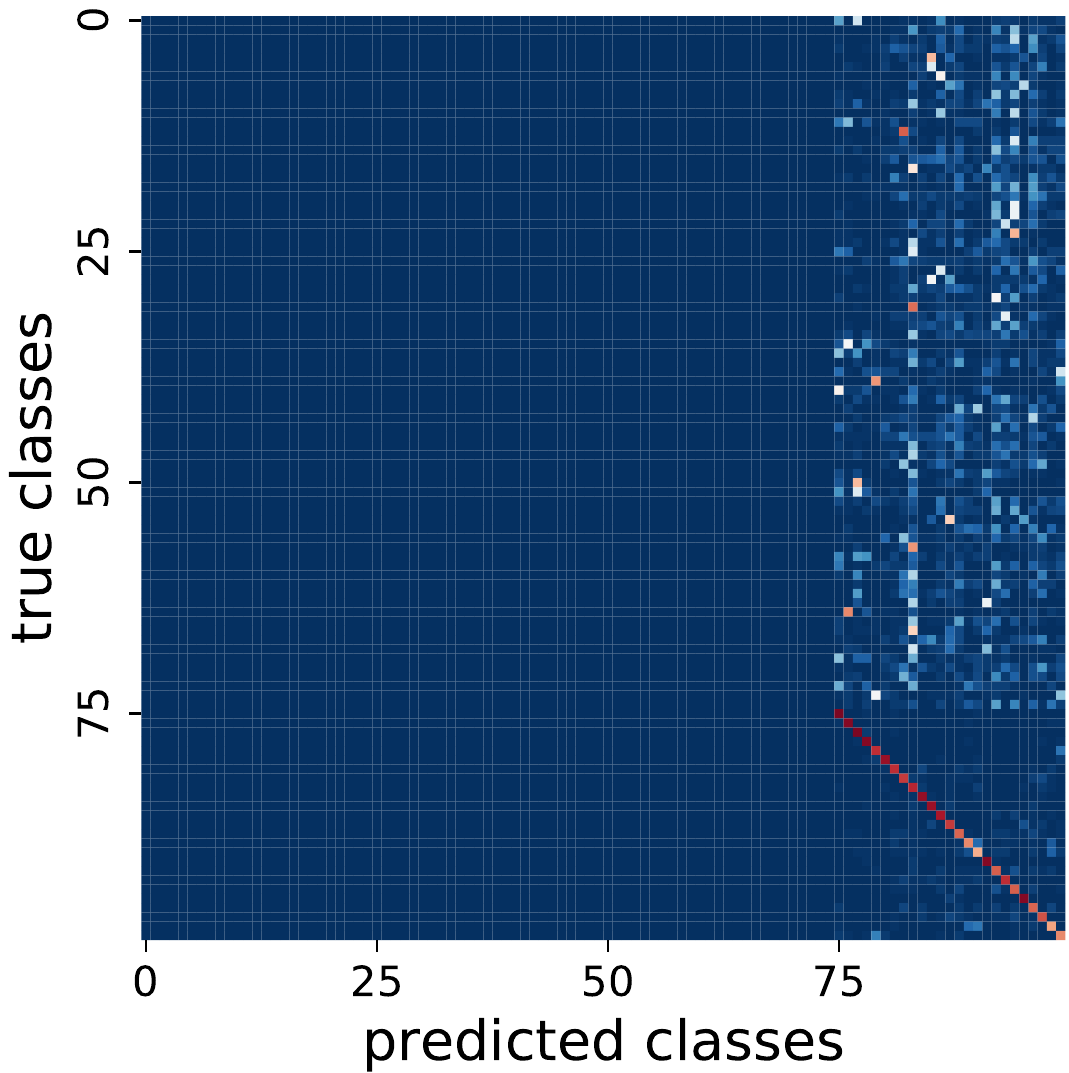}}
    \subfigure[GFR]{\label{fig:6b}\includegraphics[height=4.3cm]{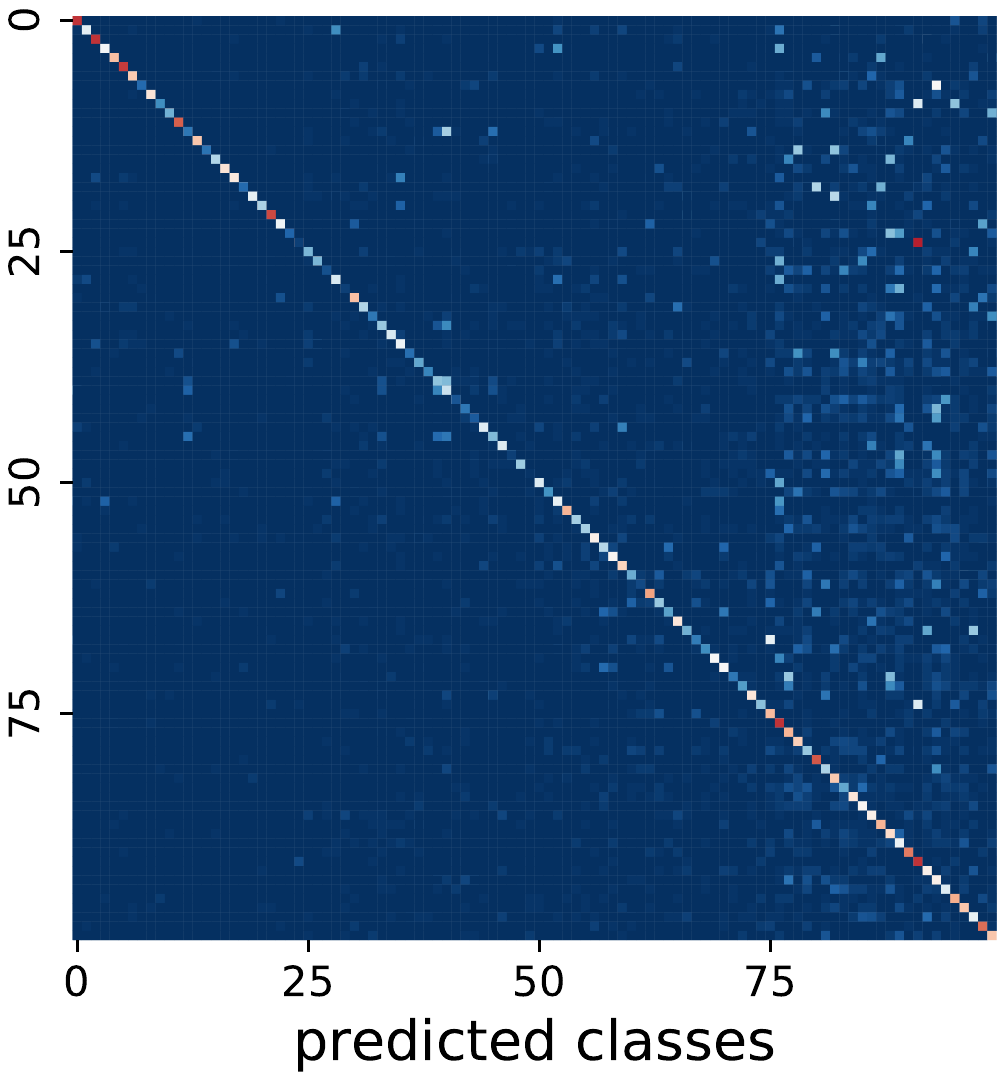}}
    \subfigure[eTag]{\label{fig:6c}\includegraphics[height=4.3cm]{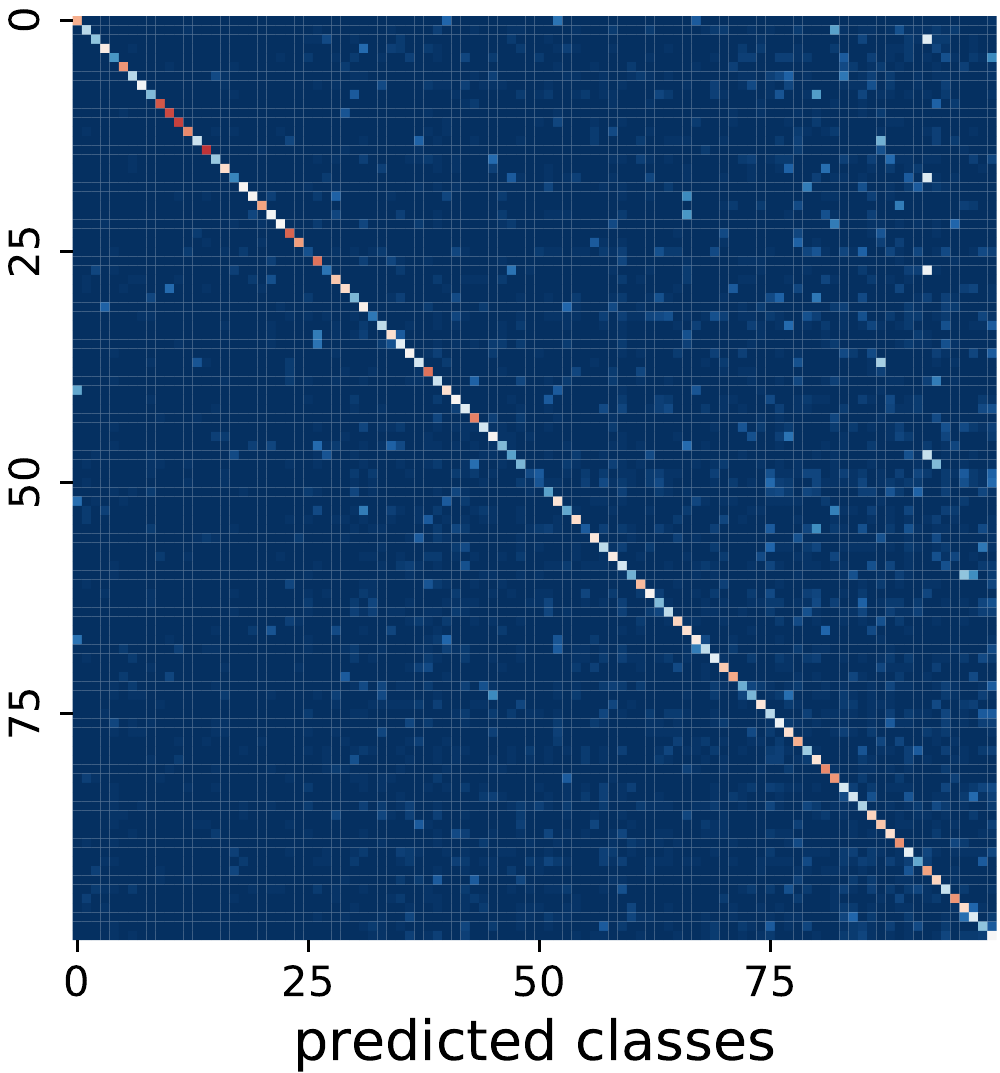}}
    \subfigure[Joint]{\label{fig:6d}\includegraphics[height=4.3cm]{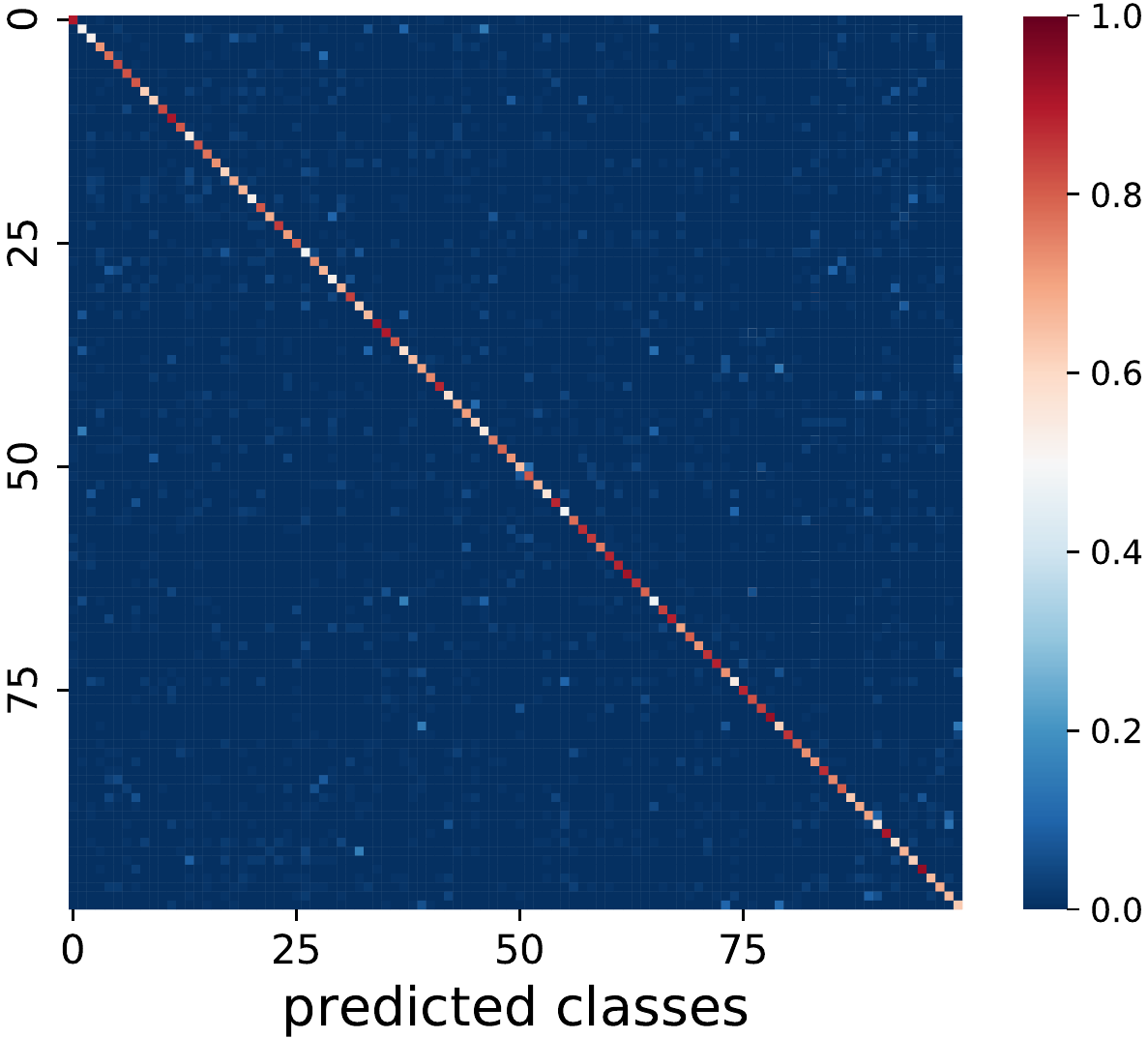}}
    \caption{Confusion matrix of (a) Fine, (b) GFR, (c) eTag, and (d) Joint.}
    \label{fig:a6} 
\end{figure*}

Besides, from Fig.\ref{fig:a6}, we find that eTag primarily improves performance by alleviating the class imbalance problem in GFR. Such a problem is majorly due to preserving no historical task samples (exemplars) or features (prototypes). As baselines, Fine suffers heavily from this problem while Joint avoids it.

\subsection{Time Consumption} \label{app:a4}
The high training time is usually a major challenge for self-supervised learning as well as training the generative models. We conduct experiments with Podnet~\citep{douillard2020podnet}], ABD~\citep{smith2021always}, and other baselines, B0, B1, B2, and B3, to evaluate the training and testing time of each component of eTag. 
As shown in Tab.\ref{tab:ablstu}, B1 and B2 have the shortest training time because they do not use a self-supervised learning task. Compared with Podnet, the increased training time of B0 is mainly from the auxiliary classifier used for embedding distillation, and this embedding distillation is not engaged in a self-supervised learning task. B3 and eTag consume the heaviest training time as they use a self-supervised learning task. It is worth noting that eTag's training time is in the acceptable range compared with the recent data-free CIL method, ABD. Moreover, the auxiliary classifier and the self-supervised learning task are used for embedding distillation only, which are not required for inference, so all baselines share almost identical testing time.

\begin{table}[!t]
  \caption{Training $+ testing$ time of baseline methods.}
  \label{tab:ablstu}
  \centering
  \resizebox{.998\linewidth}{!}{
    \begin{tabular}{cc|ccccc}
    \toprule
    PodNet & ABD & B0 & B1 & B2 & B3 & eTag \\
    \midrule
    4793.09 & 13131.21 & 6406.54 & 3050.58 & 3131.67 & 12814.07 & 12858.18 \\
    $+11.80$ & $+14.31$ & $+12.78$ & $+12.12$ & $+13.13$ & $+12.85$ & $+13.87$ \\
    \bottomrule
    \end{tabular} }
\end{table}

\begin{table}[!t]
  \caption{Incremental accuracy of various generative CIL methods.} 
  \label{tab:comwit}
  \centering
    \begin{tabular}{cccc}
    \toprule
    BI-R & GC & ABD & eTag \\
    \midrule
    34.38{\tiny$(\pm0.21)$}	& 49.55{\tiny$(\pm 0.06)$} & 50.17{\tiny$(\pm0.43)$} & 54.36{\tiny$(\pm0.22)$} \\
    \bottomrule
    \end{tabular}
\end{table}

\subsection{Compared with Sample Generative Methods} \label{app:a5}
We conduct an experiment on CIFAR-100 with ten evenly divided tasks in the same setting as GC~\citep{van2021class}. As shown in Tab.\ref{tab:comwit}, eTag achieves the best incremental accuracy because it takes task-oriented feature generation while others take original sample generation.

\end{document}